\documentclass[10pt,twocolumn,letterpaper]{article}

\usepackage{iccv}
\usepackage{times}
\usepackage{epsfig}
\usepackage{graphicx}
\usepackage{amsmath}
\usepackage{amssymb}
\usepackage{booktabs}

\usepackage[normalem]{ulem}

\usepackage[pagebackref=true,breaklinks=true,letterpaper=true,colorlinks,bookmarks=false]{hyperref}

\usepackage{colortbl}
\definecolor{Gray}{gray}{0.9}

\usepackage[capitalize]{cleveref}
\crefname{section}{Sec.}{Secs.}
\Crefname{section}{Section}{Sections}
\Crefname{table}{Table}{Tables}
\crefname{table}{Tab.}{Tabs.}

\def\name{\textsc{ACR}\xspace}

\iccvfinalcopy 

\def\iccvPaperID{46} 

\ificcvfinal\fi

\begin{document}

\title{All-pairs Consistency Learning for Weakly Supervised Semantic Segmentation}

\author{Weixuan Sun$^{1,2,3}$, Yanhao Zhang$^{5}$, Zhen Qin$^{2}$, Zheyuan Liu$^{1}$, , Lin Cheng$^{4}$, Fanyi Wang$^{5}$, \\
Yiran Zhong$^{2,3}$, Nick Barnes$^{1}$
\\
$^{1}$Australian National University,  $^{2}$OpenNLPLab, $^{3}$Shanghai AI Lab,  \\
 $^{4}$Xiamen University, $^{5}$OPPO Research Institute.
}

\maketitle
\ificcvfinal\thispagestyle{empty}\fi

\begin{abstract}
In this work, we propose a new transformer-based regularization to better localize objects for 
Weakly supervised semantic segmentation (WSSS).
In image-level WSSS, Class Activation Map (CAM) is adopted to generate object localization as pseudo segmentation labels.
To address the partial activation issue of the CAMs,
consistency regularization is employed to maintain activation intensity invariance across various image augmentations.
However, such methods ignore pair-wise relations among regions within each CAM, 
which capture context and should also be invariant across image views. 
To this end, we propose a new all-pairs consistency regularization (\name). 
Given a pair of augmented views, 
our approach regularizes the activation intensities between a pair of augmented views, while also ensuring that the affinity across regions within each view remains consistent.
We adopt vision transformers as the self-attention mechanism naturally embeds pair-wise affinity.
This enables us to simply regularize the distance between the attention matrices of augmented image pairs. 
Additionally, we introduce a novel class-wise localization method that leverages the gradients of the class token.
Our method can be seamlessly integrated into existing WSSS methods using transformers without modifying the architectures.
We evaluate our method on PASCAL VOC and MS COCO datasets.
Our method produces noticeably better class localization maps (67.3\% mIoU on PASCAL VOC \textit{train}), resulting in superior WSSS performances. 
\url{https://github.com/OpenNLPLab/ACR_WSSS}.
\end{abstract}

\begin{figure}[!t]
   \begin{center}
   {\includegraphics[width=0.9\linewidth]{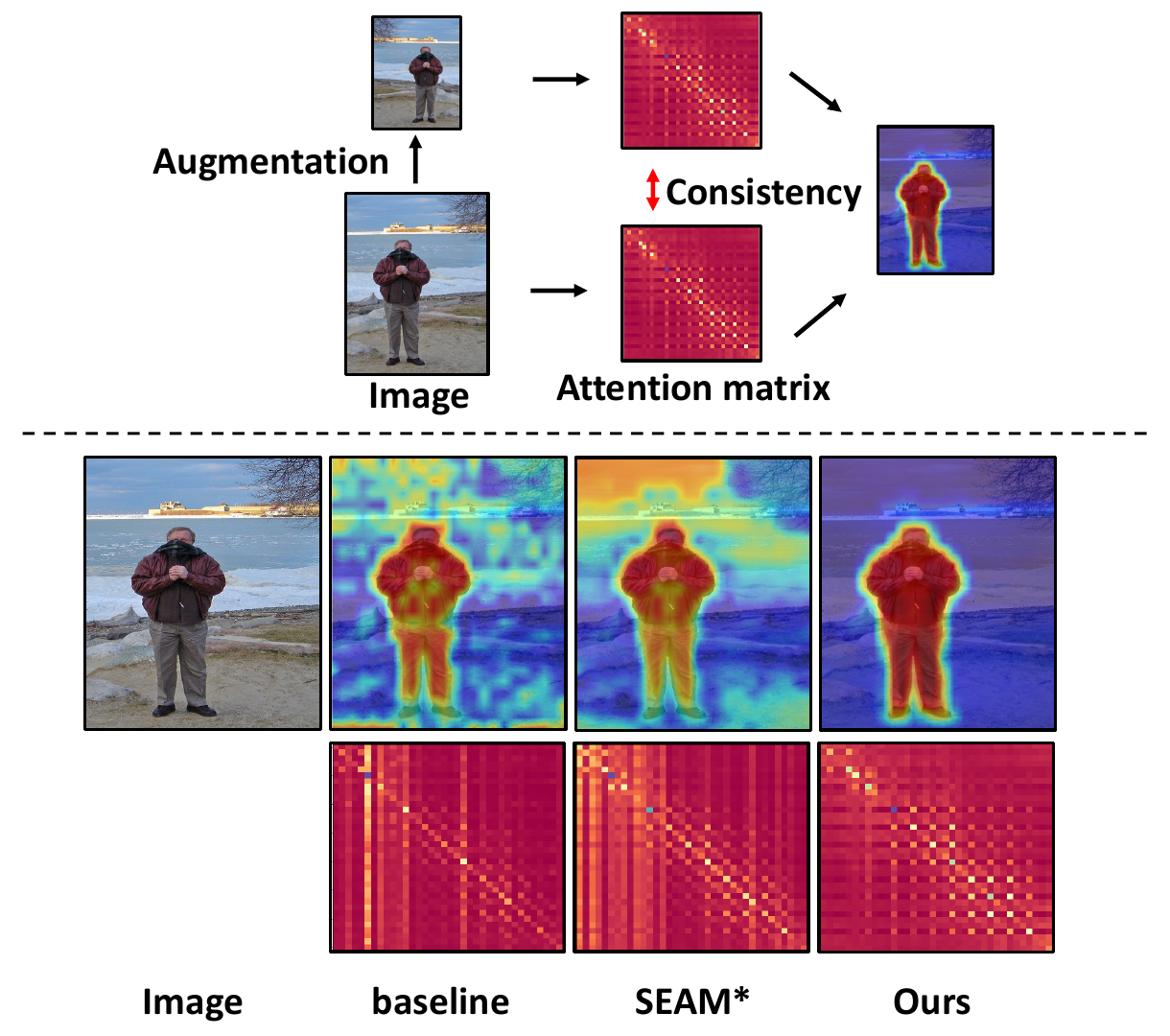}} 
   \end{center}
\caption{\textbf{Top:} conceptual illustration of the proposed \name.
Given two views of the same image by \eg, resizing \& flipping, 
we regularize the consistency between the corresponding positions of the two self-attention matrices, employing two types of invariant consistencies,
\ie 
\textit{Region affinity consistency} and \textit{Region activation consistency}.
\textbf{Bottom:} object localization and their corresponding attention matrices, all results are obtained based on vision transformer with only class labels.
Baseline: the model is trained with only classification loss. 
Other WSSS models, e.g., SEAM \cite{wang2020self}, only perform learning with activation consistency. ~SEAM*, a transformer variant of SEAM.
Ours: trained with our ACR shows the benefit of including affinity consistency.
Our approach can effectively localize targeted objects. 
}
   \label{fig:intro}
\end{figure}

\section{Introduction}
\label{section intro}
Weakly supervised semantic segmentation (WSSS) aims to relieve the laborious and expensive process of pixel-wise labeling with different types of weak labels including image-level labels \cite{huang2018weakly,ahn2018learning,fan2020cian,xu2022multi, sun2022inferring}, points \cite{bearman2016s},
scribbles \cite{vernaza2017learning,lin2016scribblesup,tang2018regularized} and bounding boxes \cite{dai2015boxsup,papandreou2015weakly,lee2021bbam,oh2021background, sun20203d}.
Image-level WSSS is particularly challenging as it uses only class labels to supervise pixel-wise predictions without any location prior.
An essential step of image-level WSSS is to obtain class-wise localization maps, i.e., seeds, which provide object localization to generate pseudo segmentation labels.
Previous WSSS methods\cite{wang2018weakly,wang2020self,chang2020weakly,Kweon_2021_ICCV,ahn2018learning,su2021context,Sun_2021_ICCV,ahn2019weakly} generally rely on Class Activation Maps (CAMs)~\cite{zhou2016learning} based on the Convolutional Neural Networks.
Although significant research has been undertaken to improve CAMs, it still suffers from incomplete and inaccurate activation.
These issues are caused by the supervision gap between the image tags and pixel-wise segmentation supervision since the classification network is indifferent to pixel-wise activation and only requires a sufficient average pooled value.

Existing work~\cite{wang2020self,Zhang_2021_ICCV,du2022weakly} uses augmentation invariant consistency to refine CAMs,
where they consider \textit{region activation consistency}
which forces the absolute class activation values to be consistent between augmented views.
Although such regularization has been demonstrated to be effective such as ~\cite{wang2020self,Zhang_2021_ICCV,du2022weakly}, activation consistency can only discover activation in novel views but non-activated regions and background noise cannot be solved through contextual relations. 
Thus, we propose to also maintain pair-wise consistency across the views, termed \textit{region affinity consistency}. 
Specifically, we look at the relations
between regions within each image and compare these relations across views.
In Fig.~\ref{fig:intro}, this implies that the relation intensities, \eg~between the person and sky, should stay invariant to augmentations between two views.
Our motivation is that affinity is a manner of context encoding and context has been demonstrated to be essential for pixel-wise predictions~\cite{Zhang_2021_ICCV,yuan2018ocnet,wang2020self}. 
Thus, every region in an image is encouraged to have the same relationships with all other regions as the augmented view, rather than simply the same value (such as
SEAM~\cite{wang2020self}). 
So both targeted and non-targeted objects are reinforced by affinity consistency.
Samples in Fig.~\ref{fig:intro} validate our motivations.
The attention matrices
of baseline and SEAM are distracted by specific tokens
(bright columns) which are not desired since an region is either targeted or non-targeted, while our method captures better object shapes 
(Diagonal grid patterns show that the targeted and non-targeted image regions are clearly distinguished).


Our method, named All-pairs Consistency Regularization (\name), uses a vision transformer to simultaneously enforce region activation consistency and region affinity consistency.
Transformer-based models have achieved great success in various tasks~\cite{dosovitskiy2020image,xie2021segformer,zhen2022cosformer,zhu2020deformable,qin2022devil,wu2021cvt,wang2021pvtv2,liu2021swin}.
As the core of the transformer, 
we find that the self-attention matrices 
can be naturally used to regularize our two consistencies without requiring additional affinity computation.
Specifically, given an image that is split into $h \times w = n$ patch tokens, an attention matrix $A\in\mathbb{R}^{(n+1)\times (n+1)}$ is generated in the self-attention module.
Its first row encodes relations between the class token and the patch tokens, 
such a class-to-patch attention can be reshaped to an $h\times w$ 
map showing potential object activation~\cite{caron2021emerging,xu2022multi,chefer2021transformer,sun2023vicinity}
for the \textit{region activation consistency}. 
Additionally, the patch-to-patch attention
$A$\texttt{[1:,1:]}$\in\mathbb{R}^{n \times n}$ encodes pair-wise relations among all pairs of patch tokens that can be used for the \textit{region affinity consistency}.
During classification training, we input the image and its augmented view into a Siamese vision transformer to obtain attention matrices for the two views respectively.
For the augmented attention matrix, we introduce a novel approach to restore the original spatial order inverting the transformation. 
Therefore we can directly regularize the corresponding positions of the attention matrices across two views to enforce the two consistencies.

The attention-based consistency is class agnostic, therefore, we cannot directly obtain a class-wise localization for the downstream WSSS task. 
Further, simply transplanting the CNN-based CAM~\cite{zhou2016learning} to transformers
relies on the output features (\ie patch tokens), 
but extensive noise is observed~\cite{xu2022multi,ru2022learning,sun2021getam}.
To this end, we
propose a new class localization generation method for vision transformers with a single class token. 
Thanks to our consistency regularization during training, the attention matrices encode rich class-wise object information. 
We find that the class-wise gradients of the class-to-patch attention $\bigtriangledown A\texttt{[0,1:]}\in\mathbb{R}^{n}$ already provide decent class-wise object localization. 
We additionally leverage the patch-to-patch attention $A\texttt{[1:,1:]}\in\mathbb{R}^{n \times n}$ to refine our class-wise localization maps and generate segmentation seeds. 
We note that our training regularization and seed generation method can be seamlessly integrated into the vision transformer networks.

To summarize, our main contributions are:
\begin{itemize}
    \item We propose All-pairs Consistency Regularization (ACR) for wsss.
    It ensures affinity consistency as well as activation consistency during the classification training, which leads to better initial seeds for wsss.
    \item We propose to leverage the self-attention structure of the vision transformer to regularize the two types of consistencies,
    which can be directly used on vision transformers without modifications. 
    To enforce the regularization, we propose a technique to re-align the spatial orders of the two views' self-attention matrices that 
     inverts the effects of a broad range of image transformations.
    \item We propose to use the gradients to generate accurate class-wise localization maps from a single class token, 
    and further refine it with the learned region affinity.
\end{itemize}

The proposed method generates significantly improved class-wise localization maps compared to all previous WSSS methods and leads to state-of-the-art performance on PASCAL VOC and MS COCO.

\begin{figure*}[!t]
   \begin{center}
   {\includegraphics[width=0.8\linewidth]{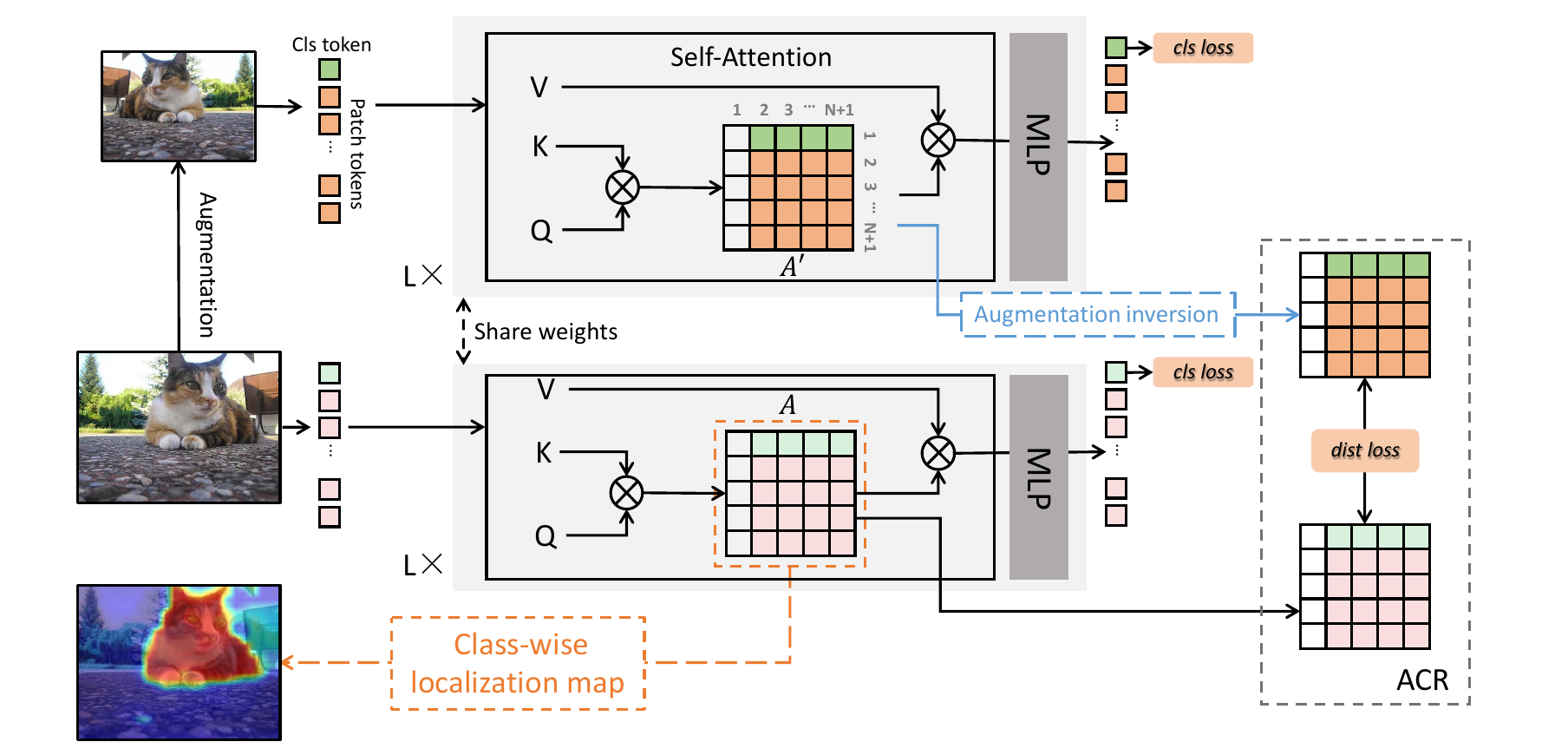}} 
   \end{center}
\vspace{-3mm}
\caption{An overview of \name. An image is augmented to a novel view then the augmented pair is input into a Siamese vision transformer (two branches share weights), consisting of $L$ successive transformer blocks. 
The class token (green) is used to make classification predictions.
In each self-attention matrix, class-to-patch attention (green) encodes region activation and patch-to-patch attention (pink) encodes region affinity.
We propose regularizing the distance between two views' self-attention matrices to enforce \name.
Our class localization map is generated using the self-attention matrix as shown in the bottom dashed orange box detailed in Fig.~\ref{fig:cam}.  
}
\vspace{-2mm}
   \label{fig:overview}
\end{figure*}

\section{Related Work}
\label{section related}
Various WSSS methods are proposed to avoid laborious pixel-wise annotation. 
The adopted weak labels include image-level labels~\cite{ahn2018learning,papandreou2015weakly,wang2020self,chang2020weakly,zhang2020reliability,yun2019cutmix,lee2021railroad,Zhang_2021_ICCV,sun2023alternative}, scribbles~\cite{lin2016scribblesup}, points~\cite{bearman2016s}, and bounding boxes~\cite{dai2015boxsup,lee2021bbam,oh2021background}.
We mainly focus on image-level methods in this review.
Existing image-level WSSS methods generally rely on CAMs~\cite{zhou2016learning} as initial seeds to generate pseudo segmentation labels. Various solutions are proposed to refine the CAMs.

\paragraph{Consistency Regularization.}
Different types of consistencies are proposed to refine the initial seeds for WSSS.
~\cite{Zhang_2021_ICCV} studies CAMs consistency from complementary patches of the same image.
~\cite{zhang2020splitting} explores the consistency between two parallel classifiers which tries to increase the distinction between the CAMs and merge the two-branch outputs to obtain complete CAMs.
Further, foreground-background contrastive~\cite{Chen_2022_CVPR,Xie_2022_CVPR} and intra-class contrastive~\cite{Sun_2021_ICCV} are proposed to refine the localization accuracy.
~\cite{sun2020mining,fan2018cian} introduce feature consistency across paired images from the same class to mine more regions.
Finally, ~\cite{du2022weakly} proposes a prototype-based metric learning methodology, that enforces feature-level consistencies in both inter-view and intra-view regularizations.
A similar method to \name is SEAM proposed in ~\cite{wang2020self}.
However, it only enforces CAM invariant consistency across augmentations but does not consider affinity consistency, i.e., the CAM values should be the same across different augmented image views. 


\paragraph{Learning Affinity Refinement.}
Pair-wise affinity is often adopted in WSSS to refine the initial seeds.
~\cite{Xu_2021_ICCV} uses an auxiliary saliency detection task to learn the affinity. 
~\cite{wang2020self,Zhang_2021_ICCV} adopts the low-level feature maps from a CNN network to generate affinity that preserves detailed context information.
~\cite{ahn2018learning,ahn2019weakly} propose to learn a network to discriminate paired pixels from the reliable seeds of CAMs. Then they use the learned network to guide random walk propagation to refine CAMs.
In the transformer era, affinity is inherently encoded in the self-attention module. 
\cite{ru2022learning} adopts reliable seeds of CAMs to directly supervise the affinity of the self-attention to capture object shapes. 
\cite{xu2022multi} adopts multiple class tokens to generate class-wise localization maps and also uses the affinity from the self-attention to refine the maps.

In summary, existing WSSS methods disregard the consistency of the affinity across views, i.e., 
In this work, we explore leveraging the self-attention mechanism to enforce such consistency.

\section{Method}
\label{section method}
In this section, we first outline the key design choices for the proposed regularization, then present our \name training framework.
Fig.~\ref{fig:overview} outlines our framework.
Our two forms of regularization 
are applied to a vision transformer~\cite{dosovitskiy2020image} without modifying the network structure.
In Section~\ref{sec:class-localization-map}, we detail our approach in obtaining the class localization maps from the network gradients.

\subsection{Overview}
We base our design on the vision transformer~\cite{dosovitskiy2020image, sun2023vicinity, wu2021cvt, wang2021pvtv2,liu2021swin}, 
as existing work~\cite{xu2022multi,ru2022learning,sun2021getam} has demonstrated that better activation is obtained. 
Compared to CNNs~\cite{wang2020self,Zhang_2021_ICCV,chang2020mixup,sun2020mining,Chen_2022_CVPR, sun2022inferring}, transformers explicitly encode region dependencies among all tokens with self-attention layers. 
Such characteristics naturally suit our need for modeling the two forms of consistencies without introducing extra modules. 
Specifically, we use class-to-patch attention to achieve region activation consistency, which sets our method apart from existing CNN-based work~\cite{wang2020self,Zhang_2021_ICCV}. 
Moreover, 
although previous work~\cite{wang2020self,Zhang_2021_ICCV} involves extra dedicated modules that model the affinity within each image, they do not use such a concept for regularizing the consistency across multiple views. We instead directly leverage the patch-to-patch attention to achieve region affinity consistency.

\subsection{Attention Consistency Regularization}
Here, we present the design of \name, with notation following ~\cite{dosovitskiy2020image}.
We split the input image into $n = h \times w$, (height by width) non-overlapping patches and flatten them to a sequence of $n$ tokens. 
A class token is inserted to form the input sequence $T \in \mathbb{R}^{(n+1)\times d} $ where $d$ is the embedding dimension. 
The class token attends to all patch tokens and is used for classification prediction.
Within each transformer block, we obtain attention matrix $A\in \mathbb{R}^{(n+1) \times (n+1)}$ by $\text{softmax}({Q {K}^{T}} / {\sqrt{d}})$~\cite{vaswani2017attention}, where $Q,K$ $\in \mathbb{R}^{(n+1)\times d}$ are the query and key matrices projected from $T$.

During classification training, we augment the input image $I$ directly to a novel view $I'$ by a randomly selected transformation.
Then we input the two views into a Siamese vision transformer to obtain two attention matrices $A$ and $A'$.
As discussed, self-attention encodes region activation and region affinity simultaneously, we calculate the distance between the two matrices to enforce our attention consistency regularization. 
To handle matrices that are not spatially equivalent after augmentations, we propose a method that rearranges the order of the tokens accordingly.
We introduce the two proposed regularization terms, as well as the token-rearranging method in detail below.

\paragraph{Region Activation Consistency} encourages the network to generate object localization that is invariant to transformations.
Consider the first row of the attention matrix $A$, we can extract the class-to-patch attention $A\texttt{[0,1:]}\in \mathbb{R}^{1 \times n}$. 
As discussed in~\cite{gao2021tscam,chefer2021generic,sun2021getam,xu2022multi,caron2021emerging},
$A\texttt{[0,1:]}$
can be reshaped and normalized to a class-agnostic objectness map $M \in \mathbb{R}^{h \times w}$ as the class token is used for classification.
Thus, given the attention matrix $A$ and its augmented view's attention matrix
$A'$,
we regularize the activation across two views by comparing the class-to-patch attention:
\begin{eqnarray}
\label{equation region activation regu}
L_\text{{act}} = \lVert A\texttt{[0,1:]} - f^{-1}(A'\texttt{[0,1:]})\rVert _{1},
\end{eqnarray}
where $f^{-1}$ is an inverse transformation to restore the spatial ordering of the tokens after the image has undergone an augmentation such as flip. 
So $f^{-1}A'$ and $A$ have the same spatial ordering of tokens,
but different values, 
since the image transformation also alters pixel ordering within each of the patches themselves, leading to altered features.
In other words, we do not invert the embeddings of the tokens but only their relative positions. 
The inversion ensures that we can regularize the corresponding positions of the two attention matrices.
The class token attends to all patches, so $n$ image patch tokens correspond to $A\texttt{[0,1:]}\in \mathbb{R}^{1 \times n}$. 
In training, we calculate the $\ell_1$ loss between corresponding areas of the two attention matrices to enforce region activation regularization.

\paragraph{Region Affinity Consistency} encourages pair-wise relations between image regions to be invariant to transformations.
Given attention matrix $A$ and its augmented view attention $A'$, and considering that $A\texttt{[1:,1:]}\in\mathbb{R}^{n \times n}$,  encodes the affinity between all patch tokens,
the affinity consistency regularization is formulated as:
\begin{eqnarray}
\label{equation region affinity regu}
L_\text{{aff}} = \lVert A\texttt{[1:,1:]} - f^{-1}(A'\texttt{[1:,1:]}) \rVert_{1}.
\end{eqnarray}
\noindent
During training, we measure $\ell_1$ loss between the corresponding pair-wise patch tokens of the two attention matrices to enforce region affinity regularization.

\paragraph{Transformation Inverse and Optimization Objective.}
Image augmentation changes the appearance and the relative positions of the patch tokens 
.
Thus, the attention matrices from two views may not be spatially equivalent, which prohibits direct distance calculation. 
To address this, we introduce a transformation to
invert the image augmentation of the attention matrix in terms of token ordering. 
Note that we only consider token ordering in this section and omit the transformation that is applied inside each image patch as we only aim to restore the original spatial information, not the embedding.
This operation is shown in the dashed blue box in Fig.~\ref{fig:overview} and
denoted as $f^{-1}$ in equation~\ref{equation region activation regu}, ~\ref{equation region affinity regu},

we present the details of the inversion in this section.


In practice, we use spatial transformations including resize, flip, and rotation.
Resize does not affect token ordering so we can simply resize the attention matrix back to the original size.
Image flip and rotation can be performed and inverted by general matrix operations.
Given a patched 
input image $X \in \mathbb{R}^{h \times w}$ 
without considering the transformation inside each patch,
flip is a permutation operation and 
rotation can be considered as a transpose followed by a flip.
So the augmented image can be formulated:
\begin{eqnarray}
\text{flip: } & X' = P_h X P_w, \\
\text{rotation: } & X' = P_h X^T P_w,
\end{eqnarray}
\noindent
where $P_h \in \mathbb{R}^{h \times h}$ and $P_w \in \mathbb{R}^{w \times w}$ are permutation matrices in x and y directions respectively. 
Let $A'$ be the attention matrix of $X'$, then the inversion of $A'$ can be written as:
\begin{eqnarray}
\label{equation inverse}
f^{-1}(A')&= C^{T}(P_w \otimes P_h^T){A'}{(P_w \otimes P_h^T)}^TC,
\end{eqnarray}
\noindent
where
$\otimes$ is Kronecker product. $C \in \mathbb{R}^{n \times n}$ is a commutation matrix for rotation and an identity matrix when flipping.
Here, we omit the class token for simplicity. 
Note that such a formulation
enables inversion of a wide range of possible image transformations that can be described by permutation matrices, though many may not be helpful augmentations.
Please refer to the supplementary material for detailed derivations and discussions.

In summary, $A$ and $f^{-1}(A')$ have the same token ordering
according to equation~\ref{equation inverse}.
Hence,
we can directly calculate the distance between the two attention matrices to apply \name.
Our optimization objective is the combination of the two-view classification and the consistency losses: 

\begin{equation}
L = L_\text{{cls}} + \alpha L_\text{{act}} + \beta L_\text{{aff}}.
\end{equation}
\noindent

\noindent where $\alpha$, $\beta$ are hyperparameters.

\subsection{Gradient-based Transformer Class Localization Map}
\label{sec:class-localization-map}
At test time, the object activation provided by the class-to-patch attention $A\texttt{[0,1:]}\in \mathbb{R}^{1 \times n}$ is class-agnostic~\cite{caron2021emerging,xu2022multi}.
To obtain class-wise localizations for the downstream WSSS task, a naive solution is to directly transplant the CAM~\cite{zhou2016learning} method into the transformer, by using the average pooled patch tokens instead of the class token to produce classification predictions. 
However, in line with existing works~\cite{sun2021getam,ru2022learning,xu2022multi}, we find that this achieves poor results (Table~\ref{table ablation cam generation}).
Another approach~\cite{xu2022multi} uses multiple class-specific tokens to generate class-wise seeds. However, this requires modifying the transformer architecture and
computational complexity grows with the number of classes.
Inspired by recent transformer interpretability work~\cite{chefer2021generic,chefer2021transformer}, we introduce a gradient-based approach.
Different from~\cite{chefer2021generic,chefer2021transformer} which incorporate gradients with the attention values or the network relevances~\cite{bach2015pixel}, 
we empirically find that the gradients can directly provide accurate localization information and construct our gradient-based transformer class localization methods.

\begin{figure}[t]
   \begin{center}
   \includegraphics[width=0.9\linewidth]{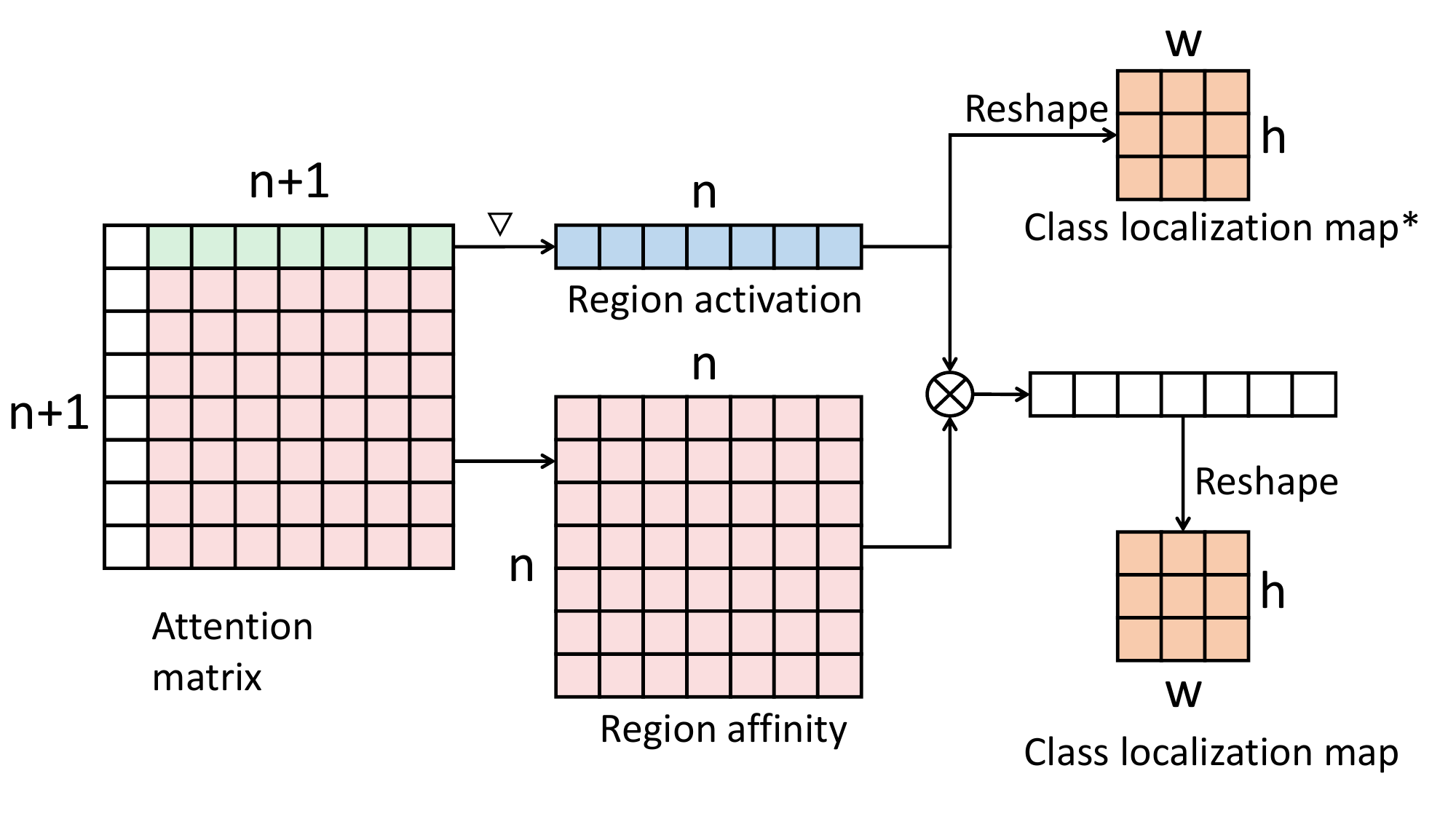}
   \end{center}
   \vspace{-3mm}
\caption{An overview of our class-wise localization map generation framework. We use the gradients of the class-to-patch attention (the blue vector) to generate a class localization map*. Further, we use the learned region affinity (the pink block) to refine the class localization map. Sample visualizations of the learned region affinity are shown in Fig.~\ref{fig:aff mat}.}
\vspace{-3mm}
   \label{fig:cam}
\end{figure}

In Fig.~\ref{fig:cam},
given the class-to-patch attention matrix $A\texttt{[0,1:]}\in \mathbb{R}^{1 \times n}$ (the green vector) and target class $c$, 
we calculate gradients by back-propagating the classification score $y^{c}$,
formulated as $\bigtriangledown A^{c}\texttt{[0,1:]}:= \partial{y^{c}} / ( \partial{A\texttt{[0,1:]}} )$ (the blue vector).
Intuitively, $\bigtriangledown A^{c}\texttt{[0,1:]}$, \ie class-wise gradients of the class-to-patch attention, represent patch tokens' contributions to the final classification scores.
Then we remove negative values and reshape it to $h \times w$ to obtain the class localization map* shown in Fig.~\ref{fig:cam}.
We empirically find that averaging the multi-layer gradients performs well.
Given a transformer with $l$ successive layers, 
the localization map for class $c$ is defined as: 
\begin{equation}
\label{equation air_0}
M^{c} = \frac{1}{l}\sum_{i}^{l}{\bigtriangledown A^{c}_{i}\texttt{[0,1:]}}.
\end{equation}
\noindent
\paragraph{Affinity Refinement.} 
Inspired by~\cite{wang2020self,xu2022multi}, we further adopt the learned patch-wise affinity $A\texttt{[1:,1:]}\in\mathbb{R}^{n \times n}$ (the pink matrix) to refine the activation maps as shown in Fig~\ref{fig:cam}. 
Thanks to our Region affinity consistency, better context is encoded in self-attention modules. 
We visualize the patch-wise affinity in Fig.~\ref{fig:aff mat}.
Note that the baseline model is trained with only classification loss without our regularization, 
the generated affinity (the third column) is distracted by specific patch tokens, leading to noisy seeds (the second column). 
Our affinity (the fifth column) can capture better object contexts and generate integral localization.
Formally, our class localization map for class $c$ is defined as:
\begin{eqnarray}
\small
\label{equation air_1}
M^{c} = \left( \frac{1}{l}\sum_{i}^{l}{\bigtriangledown A^{c}_{i}\texttt{[0,1:]}} \right) \times \left( \frac{1}{l}\sum_{i}^{l}A_{i}\texttt{[1:,1:]} \right).
\end{eqnarray}
\noindent
Then, $M^{c}\in \mathbb{R}^{1 \times n}$ can be reshaped to $h \times w$ and normalized to obtain our final class localization maps, i.e., initial seeds. 

\begin{figure*}[!t]
   \begin{center}
   {\includegraphics[width=0.8\linewidth, height=6.5cm]{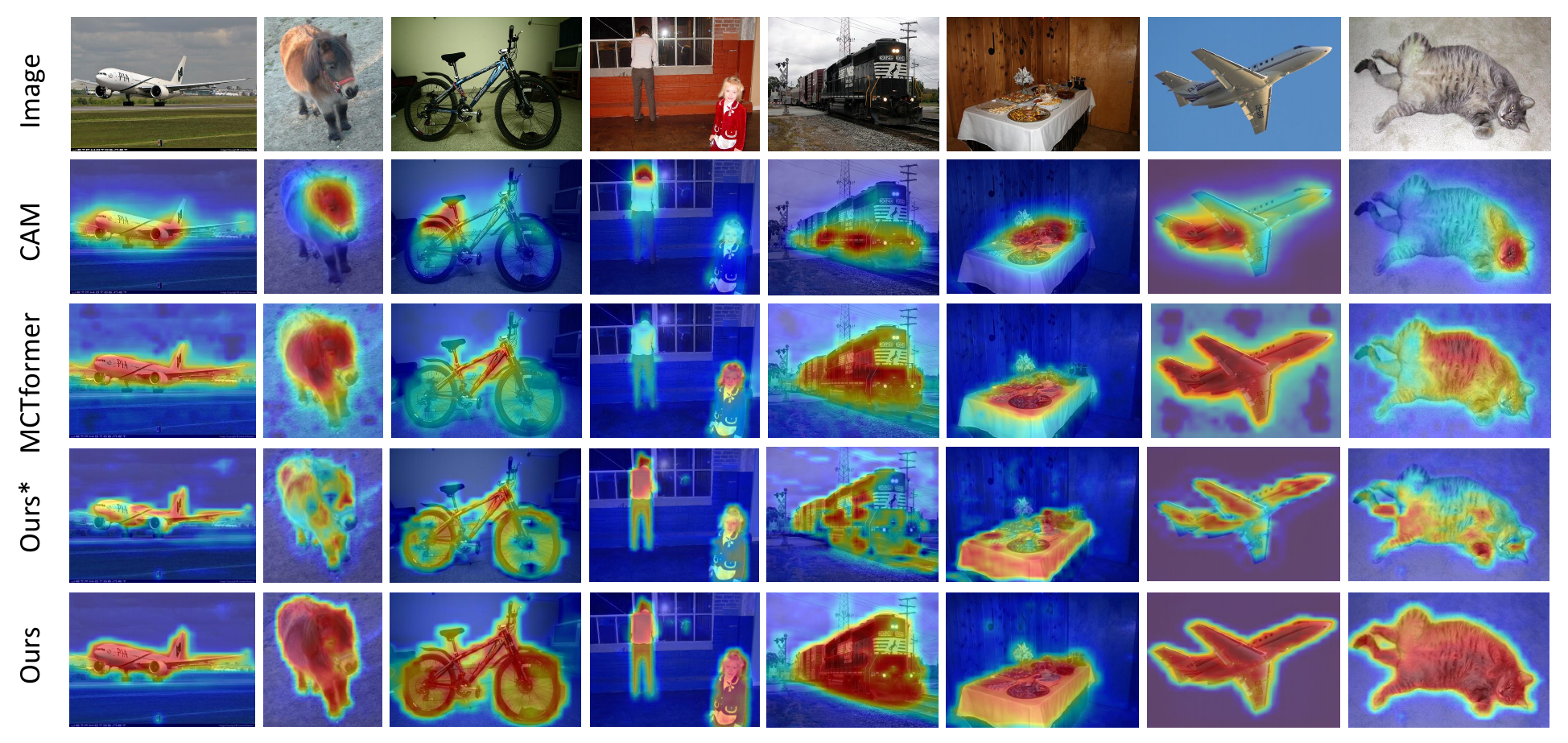}} 
   \end{center}
\caption{Visualization samples of the class localization maps of different methods. CAM: Class Activation Methods~\cite{zhou2016learning}. MCTformer: class localization maps of~\cite{xu2022multi} which also adopt transformer attention refinement. Ours*: our class localization maps without affinity refinement. Ours: our final class localization maps with affinity refinement.}
   \label{fig:cam vis}
\end{figure*}

Our class-wise localization maps provide accurate and dense object coverage. The reasons are two-fold.
First, the region activation consistency encourages the class token to attend to accurate object localization as shown in row 4 of Fig.~\ref{fig:cam vis}.
Second, affinity consistency regularization encourages the network to capture precise pair-wise affinity, such affinity propagates the localized pixels to cover comprehensive object regions, as shown in row 5 of Fig.~\ref{fig:cam vis}.



\begin{figure}[!t]
   \begin{center}
   {\includegraphics[width=1\linewidth]{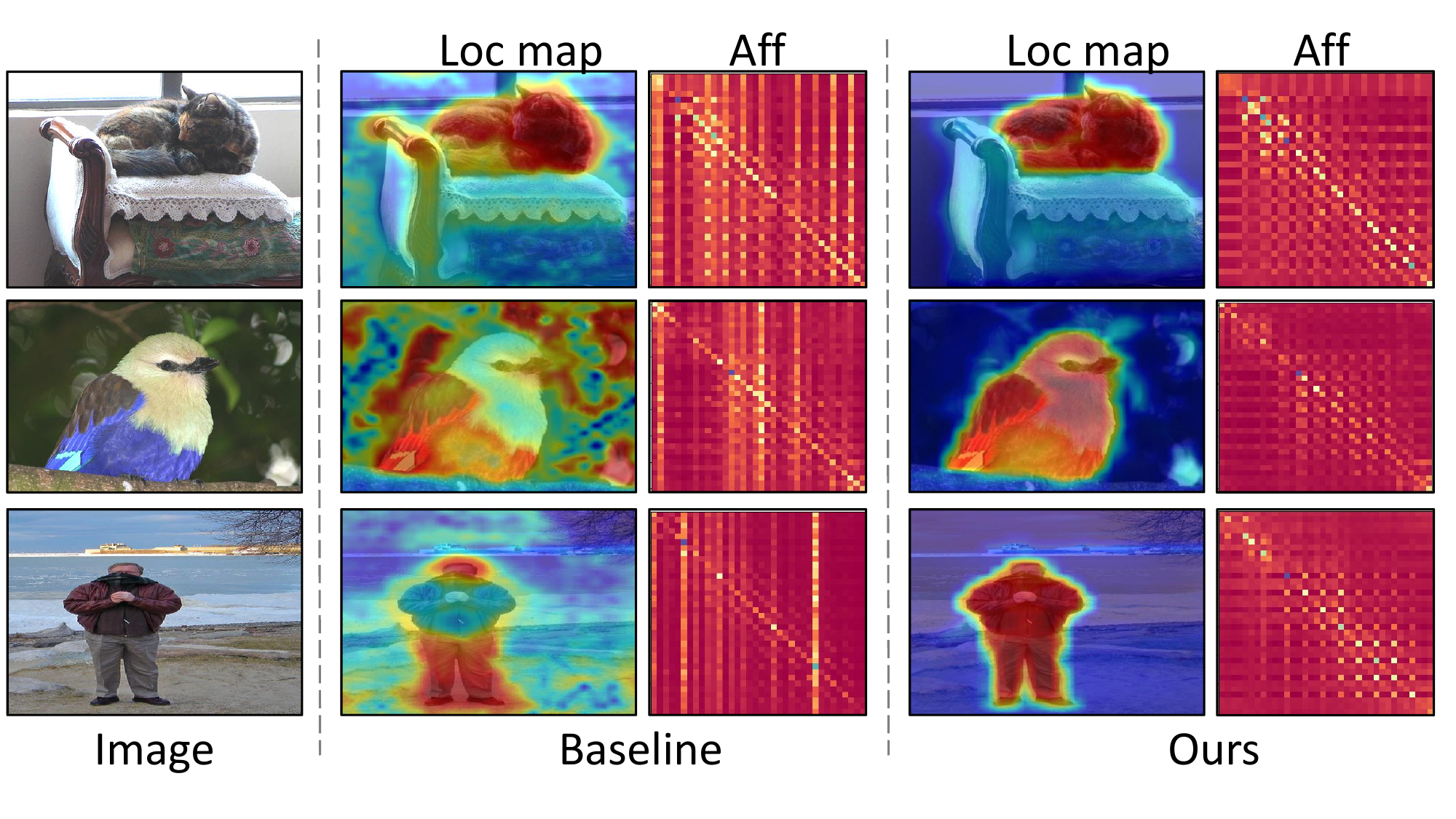}} 
   \end{center}
   \vspace{-3mm}
\caption{Class localization maps (Loc map) and pair-wise affinity of patch tokens (Aff). Our method can capture better context encoding and generate accurate localization maps. 
Baseline: the model is trained with only classification loss.
The attention matrices are down-sampled for readability.}
\vspace{-3mm}
   \label{fig:aff mat}
\end{figure}

\begin{figure}[!t]
   \begin{center}
   {\includegraphics[width=1\linewidth]{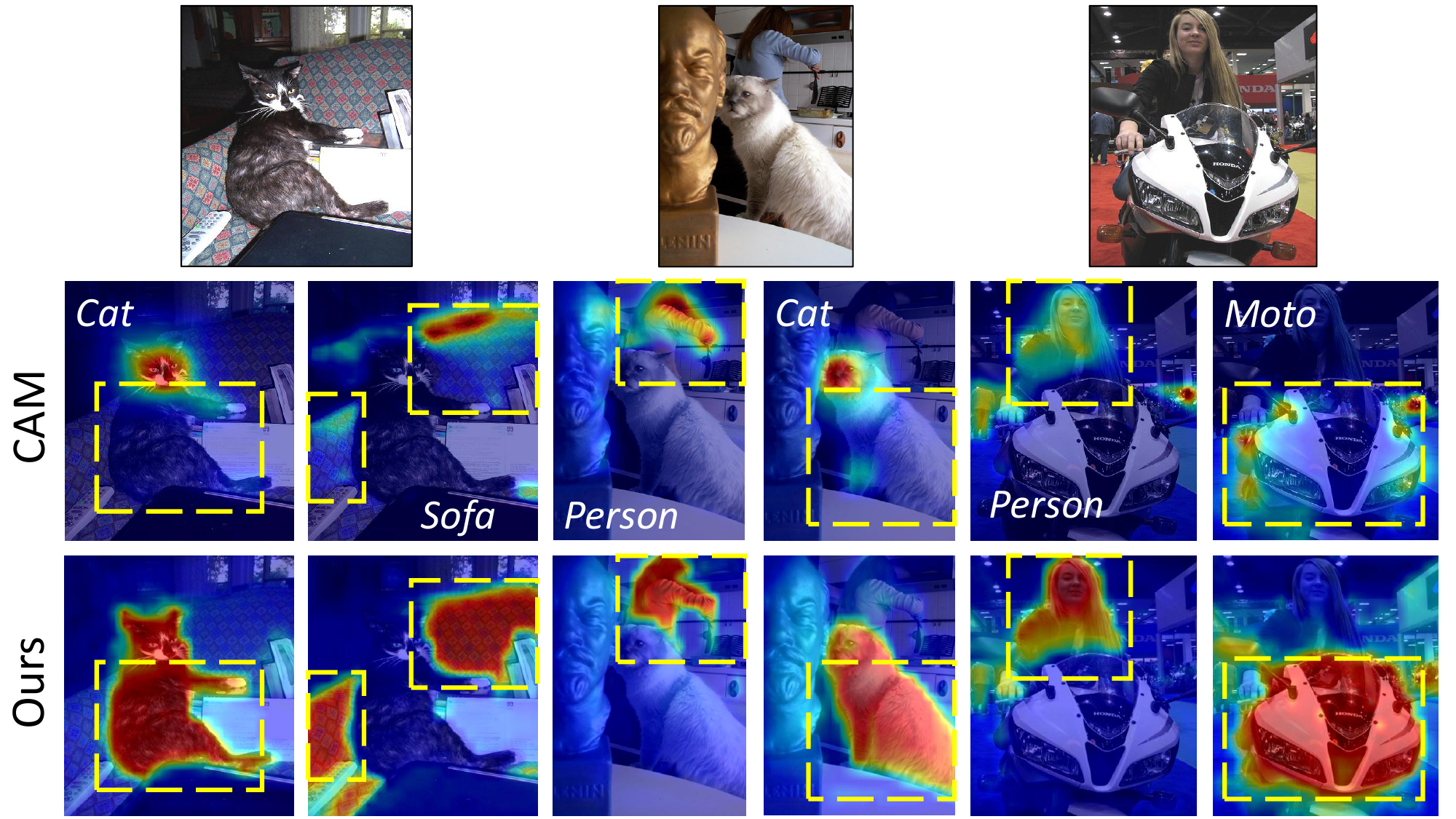}} 
   \end{center}
\caption{Visualization samples of our class localization maps with multiple classes. \name can discriminate accurate boundaries between connected objects and localize complete shapes.}
   \label{fig:cam edge}
\end{figure}

\section{Experiments}
\label{section exp}
\subsection{Experimental Settings}
\paragraph{Datasets.}
We evaluate our method on the PASCAL VOC~\cite{everingham2010pascal} and MS COCO~\cite{lin2014microsoft} datasets.
The official
PASCAL VOC has 20 objects classes and one background class,
with 1,446 training, 1,449 validation, and 1,456 testing images. Following common practice in WSSS, we use an augmented training set consisting of 10,582 images with annotations from~\cite{hariharan2011semantic}.
MS COCO 2014 
is much more challenging than PASCAL VOC. It
contains 81 classes including background with 80k training and 40k validation images.

\paragraph{Implementation Details.}
We adopt ViT-hybrid-B~\cite{dosovitskiy2020image}.
Training images are resized and cropped to $384 \times 384$.
For semantic segmentation, following previous WSSS methods~\cite{xu2022multi,ahn2018learning,ahn2019weakly,Kweon_2021_ICCV}, we use DeepLabV2~\cite{chen2017deeplab} with a ResNet101~\cite{he2016deep} backbone as the segmentation model. 
During segmentation inference, we use multi-scale testing and adopt CRFs~\cite{krahenbuhl2011efficient} for post-processing. 
Detailed implementation details are presented in the supplementary material.

\begin{table}[t!]
\caption{Performance comparison of WSSS methods on MS COCO. w/ saliency: the method adopts extra saliency information.
Best number is in bold.}
\centering\scalebox{.85}{
\setlength{\tabcolsep}{3.2mm}
\begin{tabular}{lccc}
\toprule
Methods & Venue & w/ saliency  & Val  \\
\hline
AuxSegNet \cite{Xu_2021_ICCV}& ICCV2021 & \checkmark & 33.9 \\
EPS~\cite{fan2020employing} &CVPR2022& \checkmark &  35.7 \\
L2G~\cite{Jiang_2022_CVPR}&CVPR2022 & \checkmark & 44.2 \\
\hline
Wang et al.~\cite{wang2018weakly}&IJCV2020 & & 27.7 \\
Ru et al.~\cite{ru2022learning}&CVPR2022 &  & 38.9  \\
SEAM~\cite{wang2020self}&CVPR2020 && 31.9 \\
CONTA~\cite{zhang2020causal}&NeurIPS2020 && 32.8  \\
CDA~\cite{su2021context}&ICCV2021 & & 33.2  \\
Ru et al.~\cite{ru2022weakly} & IJCV2022 & & 36.2 \\
URN~\cite{li2021uncertainty}&AAAI2022  && 41.5 \\
MCTformer~\cite{xu2022multi} &CVPR2022 &  & 42.0 \\
ESOL~\cite{li2022expansion} & NeurIPS2022 & &42.6 \\
SIPE~\cite{Chen_2022_CVPR}&CVPR2022 & & 43.6 \\
RIB~\cite{lee2021reducing}&NeurIPS2020 & & 43.8 \\
\textbf{\name}   & &   & \textbf{45.0}          \\
\hline
\end{tabular}
}

\label{table coco}
\end{table}

\begin{table}[t!]
\caption{Performances of the initial Seeds and pseudo segmentation labels on PASCAL VOC \textit{train} set. 
\textit{(s)}: methods that rely on saliency to generate seeds.
ACR*: our localization maps without affinity refinement.
Our seeds outperform previous non-salient methods by a significant margin.
}
\centering\scalebox{.85}{
\setlength{\tabcolsep}{1.8mm}
\begin{tabular}{lc cc}
\toprule
Methods & Seed & w/ saliency & Pseudo \\
\hline
EDAM (CVPR2021)~\cite{wu2021embedded} & 52.8 & \checkmark & 68.1 \\
ReCAM (CVPR2022) \cite{chen2022class}  & 54.8  & \checkmark & 70.9 \\
L2G (CVPR2022)~\cite{Jiang_2022_CVPR} & 56.2  & \checkmark &71.9\\
EPS (ECCV2020)~\cite{fan2020employing}  & 69.4 \textit{(s)}  & \checkmark & 71.6 \\ 
Du et al.(CVPR2022)~\cite{du2022weakly} & \textbf{70.5} \textit{(s)} & \checkmark & \textbf{73.3}  \\
\hline
PSA (CVPR2018)~\cite{ahn2018learning} &  48.0 &  & 61.0  \\
SEAM (CVPR2020)~\cite{wang2020self} & 55.4 & & 63.6 \\
CDA (ICCV2021)~\cite{su2021context} & 55.4 & & 67.7 \\
AdvCAM (CVPR2021)~\cite{lee2021anti} & 55.6  & & 68.0 \\ 
CPN (ICCV2021)~\cite{Zhang_2021_ICCV} & 57.4 & & 67.8 \\
Ru et al. (CVPR2022)~\cite{ru2022learning} & -- & & 68.7 \\
SIPE (CVPR2022)~\cite{Chen_2022_CVPR} & 58.6 &  & -- \\
Du et al.(CVPR2022)~\cite{du2022weakly} & 61.5  &  & 69.2 \\
MCTformer (CVPR2022)~\cite{xu2022multi}  & 61.7&  &69.1 \\

ACR* & 59.4  & & -- \\
\textbf{\name}   & \textbf{67.3} &   &     \textbf{70.8}            \\
\hline
\end{tabular}
}
\vspace{-3mm}
\label{table cam and pseudo}
\end{table}

\subsection{Comparison with State-of-the-art}

\subsubsection{MS COCO}
Table~\ref{table coco} shows segmentation results on MS COCO. 
We achieve a segmentation mIoU of 45\%, which surpasses existing methods with a clear margin.
Notably, this result does not rely on any extra saliency information but outperforms all previous WSSS methods including the ones with saliency.
MS COCO is a bigger dataset with more semantic classes and complex images that include multiple objects. This result indicates that saliency may hinder WSSS approaches' ability to scale to complex scenes, 
hence we do not incorporate saliency into our approach.
Our result demonstrates that \name is able to generate reliable class localization maps in challenging scenes. 
We report per-class results of MS COCO in the supplementary material.

\subsubsection{PASCAL VOC}
\begin{figure}[!t]
   \begin{center}
   {\includegraphics[width=1\linewidth]{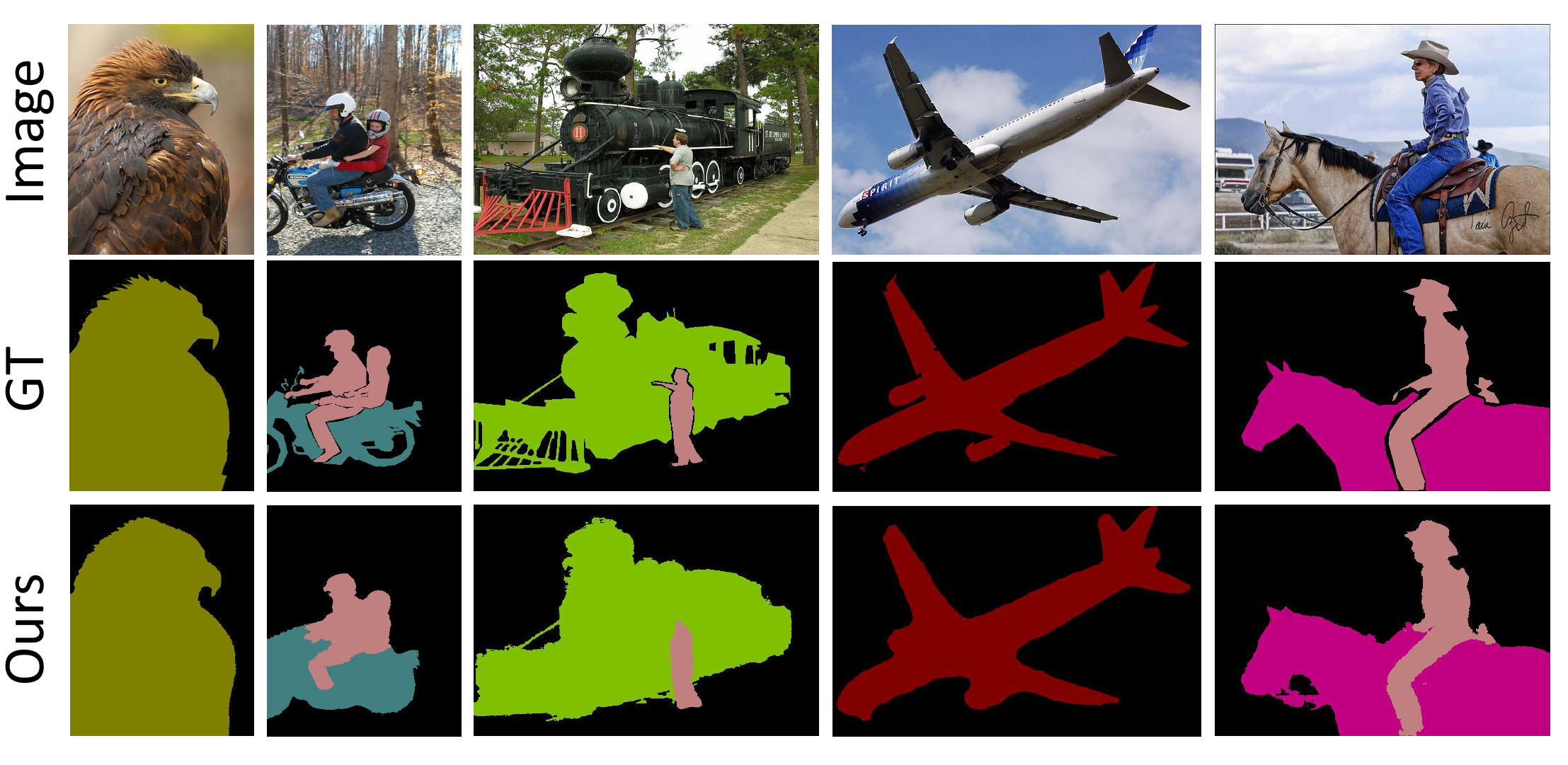}} 
   \end{center}
\caption{Segmentation results on the PASCAL VOC \textit{val} set.}
   \label{fig:seg}
\end{figure}

\paragraph{Seed Performance.}
We report mIoU for the class localization maps in Table~\ref{table cam and pseudo}, including the performances with and without affinity refinement.
As shown, without affinity refinement, ACR* still outperforms most existing non-salient methods (59.4\% mIoU).
Our \name achieves significantly improved initial seeds, which shows the efficacy of the proposed \name.  
Without the assistance of saliency, 
previous best~\cite{xu2022multi} also adopts transformer affinity to refine the seed, \name outperforms it by 5.2\%.
We show qualitative results in Fig.~\ref{fig:cam vis}. 
Further, Fig.~\ref{fig:cam edge} shows seeds on complex scenes with multiple objects, \name learns precise affinity to facilitate complete object shapes with precise boundaries.\\
\paragraph{Pseudo Label Performance.}
The last column of Table~\ref{table cam and pseudo} shows the pseudo segmentation label performances.
Following common practice, we adopt PSA~\cite{ahn2018learning} to process the activation maps (seed) into pixel-wise pseudo segmentation labels. 
We empirically find that PSA is easily affected by false positive samples, i.e., over-activation.
To avoid over-activation, we use ACR* to train the PSA network. Then, the trained PSA network will refine the \name seeds (67.3\%) into pseudo labels. 
As shown, our method achieves notably improved pseudo labels.\\
\paragraph{Semantic Segmentation Performance.}
Table~\ref{table pascal seg} shows semantic segmentation results on PASCAL VOC.
\name achieves competitive results of 71.2\% and 70.9\% on \textit{val} and \textit{test} sets respectively, which outperform previous non-salient methods.
Fig.~\ref{fig:seg} shows that the segmentation model trained
with our pseudo labels can produce accurate and complete predictions.
We report per-class results of PASCAL VOC in the supplementary material.


\begin{table}[t!]
\caption{Performance comparison of WSSS methods on PASCAL VOC 2012 \textit{val} and \textit{test} sets. 
w/ saliency: the method adopts extra saliency information. Best numbers are in bold.}
\centering\scalebox{.85}{
\setlength{\tabcolsep}{2.4mm}
\begin{tabular}{lcccc}
\toprule
Methods & Venue & w/ saliency  & Val & Test  \\
\hline
NSRM~\cite{yao2021non}    & CVPR2021 & \checkmark   & 70.4    &  70.2\\
EDAM~\cite{wu2021embedded} &CVPR2021 &\checkmark & 70.9 & 70.6 \\
EPS~\cite{lee2021railroad} &CVPR2021& \checkmark & 71.0 & 71.8 \\
DRS~\cite{kim2021discriminative} &AAAI2021& \checkmark & 71.2 & 71.4 \\
L2G~\cite{Jiang_2022_CVPR} &CVPR2022& \checkmark & 72.1 & 71.7 \\
Du et al. ~\cite{du2022weakly}&CVPR2022 & \checkmark & \textbf{72.6} & \textbf{73.6} \\
\hline
PSA~\cite{ahn2018learning} & CVPR2018&  &61.7  & 63.7  \\
SEAM~\cite{wang2020self} &CVPR2020 &  & 64.5 & 65.7 \\
CDA~\cite{su2021context}&ICCV2021 & & 66.1 & 66.8 \\
ECS-Net~\cite{Sun_2021_ICCV}&ICCV2021 &&66.6 & 67.6 \\
Du et al. ~\cite{du2022weakly}&CVPR2022 & &67.7 & 67.4\\
CPN~\cite{Zhang_2021_ICCV}&ICCV2021  &    & 67.8   &  68.5\\
AdvCAM~\cite{lee2021anti}&CVPR2021 & & 68.1 & 68.0 \\  
Kweon et al.~\cite{Kweon_2021_ICCV}&ICCV2021 && 68.4 &68.2 \\
ReCAM~\cite{chen2022class}&CVPR2022  &   &  68.5 & 68.4 \\
SIPE~\cite{Chen_2022_CVPR}& CVPR2022 & & 68.8 & 69.7 \\
URN~\cite{li2021uncertainty}&AAAI2022 & & 69.5 & 69.7 \\
ESOL~\cite{li2022expansion} & NeurIPS2022 & & 69.9 & 69.3 \\
PMM~\cite{Li_2021_ICCV}&ICCV2021  &    & 70.0    &  70.5\\
MCTformer\protect\footnotemark~\cite{xu2022multi} &CVPR2022  &  & 70.6 & 70.3 \\
VWL-L~\cite{ru2022weakly}&IJCV2022  &   & 70.6 & 70.7 \\
Lee et al.~\cite{lee2022weakly} & CVPR2022 & & 70.7 & 70.1 \\
\textbf{\name}   &   &  &\textbf{71.2} &  \textbf{70.9}          \\
\hline
\end{tabular}}
\label{table pascal seg}
\end{table}

\footnotetext{Xu et al.~\cite{xu2022multi} report 71.9 (\textit{val}) and 71.7(\textit{test}), but we are unable to reproduce these results with their provided code and seeds. We instead report our reproduced performances using their official implementation at \nolinkurl{https://github.com/xulianuwa/MCTformer}.}

\subsection{Ablation Studies}
\paragraph{Effectiveness of \name.}
We propose to simultaneously regularize region activation and region affinity during the classification training. We ablate the two regularization terms in Table~\ref{table ablate regu}. 
First, we observe that region affinity can significantly improve seed quality even in the baseline, which validates the contextual encoding ability of the vision transformer. 
By introducing the two regularization terms, we observe that they contribute noticeable improvements to the performance respectively.
We achieve superior results with both regularization terms, leading to an overall
15.8\% mIoU increase over the vanilla transformer baseline (51.1\%)
, which demonstrates the effectiveness of \name.

\begin{table}[t!]
\caption{Ablation analysis of the two proposed consistency regularization. Act Regu: region activation consistency regularization. Aff Regu: region affinity consistency regularization.
aff: whether to use affinity refinement during seed generation.
}
\centering\scalebox{.85}{
\setlength{\tabcolsep}{3.5mm}
\begin{tabular}{cc | cc}
\hline
Act Regu & Aff Regu & w/o aff 
&  w/ aff \\ \hline 
 & & 51.1 & 57.7  \\
 \checkmark & & 54.8 & 63.5 \\
 & \checkmark & 55.4 & 64.9 \\
 \checkmark&\checkmark& \textbf{59.4} & \textbf{67.3} \\
\hline
\end{tabular}}
\vspace{-3mm}
\label{table ablate regu}
\end{table}

\begin{table}[t!]
\caption{Analysis of different seeds generation methods.}
\centering\scalebox{.85}{
\setlength{\tabcolsep}{16mm}
\begin{tabular}{l r}
\toprule
Methods & Seed \\ \hline
CAM~\cite{zhou2016learning} & 44.0 \\
TS-CAM~\cite{gao2021tscam} & 40.1 \\
Grad-CAM~\cite{selvaraju2016grad} & 50.5  \\
Generic~\cite{chefer2021generic} & 52.5 \\
Generic~\cite{chefer2021generic} + aff  & 59.6 \\
\name  & \textbf{67.3} \\
\hline
\end{tabular}}
\vspace{-2mm}
\label{table ablation cam generation}
\end{table}

\paragraph{Different Seeds Generation Methods.}
\label{sec:seed_ablation}
To fully utilize the structure of the vision transformer, we integrate the class-to-patch gradients with region affinity to generate class localization maps as seeds.
In Table~\ref{table ablation cam generation}, we ablate different seeds generation methods with models trained using our \name.
We first integrate the conventional CAM~\cite{zhou2016learning} method into the vision transformer, which achieves only 44\%,
potentially because the context aggregated in the class token is not used, and the global receptive field may spread noise.
We further test various network visualization methods including TS-CAM~\cite{gao2021tscam}, Grad-CAM~\cite{selvaraju2016grad}, and Generic~\cite{chefer2021generic}. 
Notably, we refine the outputs of Generic~\cite{chefer2021generic} and observe a performance boost, which shows that the region affinity refinement can also be integrated with other methods for a performance increase. 
Ultimately, \name achieves the best result, demonstrating the effectiveness of our seed generation method.

\paragraph{Different Vision Transformer Backbones.}
in Table~\ref{table ablation backbone}, we compare \name with the previous best 
localization maps 
generated by MCTformer~\cite{xu2022multi} using the same vision transformer backbone, i.e., Deit-S~\cite{touvron2021training}, which has substantially fewer parameters and lower complexity compared to ViT-hybrid-B. 
Compared to \name, 
MCTformer produces better localization maps without affinity refinement (58.2 vs 56.8) since it uses multiple class tokens which require more complexity, while we only rely on a single one. 
However, our model is more benefited with the affinity refinement (61.7 vs 63.4). This is because our \name learns better pair-wise affinity which leads to more integral object localization. 
Moreover, we integrate our \name during the MCTformer training
(Table~\ref{table ablation backbone}: MCTformer + ACR).
As shown, ACR improves MCTformer by 2.2 mIoU without affinity and 0.7 mIoU with affinity.
In summary, 
it demonstrates that \name can work with different transformer backbones and existing transformer-based WSSS methods as well.



\begin{figure}[!t]
   \begin{center}
   {\includegraphics[width=0.8\linewidth]{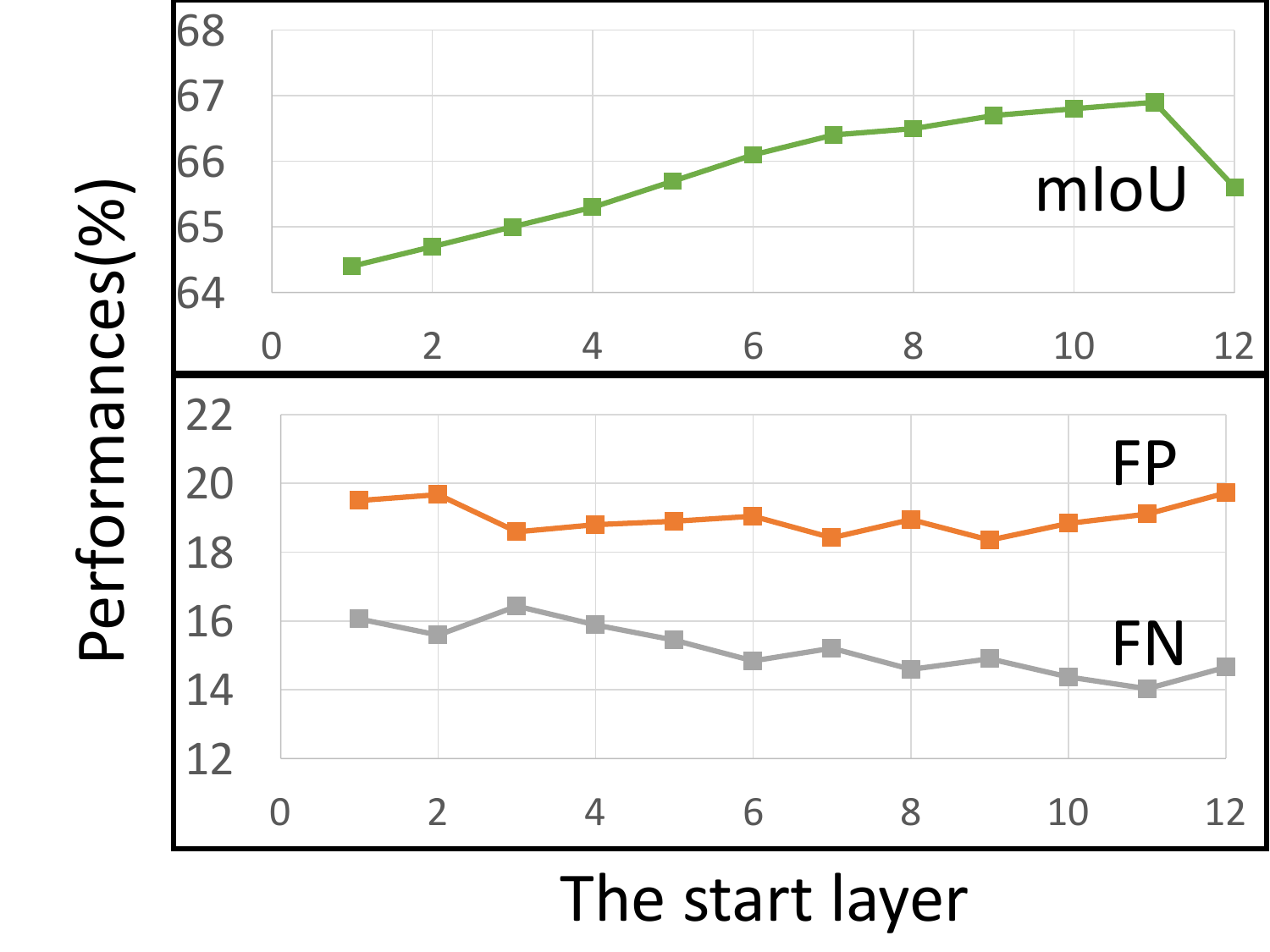}} 
   \end{center}
\caption{Performance in mIoU (\%), false positive (FP), and false negative (FN) of the initial seeds generated by averaging over transformer layers. The horizontal axis represents which layer we start to obtain seeds.}
   \label{fig:layers}
\end{figure}

\paragraph{Different Layers of CAM generation.}
We obtain the class localization maps by averaging the outputs of successive transformer layers. 
Following ~\cite{wang2020self}, 
we report mIoU, false positive (FP) and false negative (FN) of the localization maps when we fuse from different layers. 
FP indicates over-activation and FN indicates under-activation. 
As shown in Fig.~\ref{fig:layers},
the mIoU tends to increase and FN tends to decrease when reducing the number of layers used, and both values are saturate when only the last two layers are involved. 
This indicates that early layers may contain unhelpful low-level noise, and with only the last two layers, we can obtain the best object completeness. 
Further, our seeds are generally over-activated as the FP is consistently higher than the FN. It indicates that the incompleteness issue is effectively mitigated by \name.
However, current pseudo generation methods~\cite{ahn2018learning,ahn2019weakly} are designed for under-activated seeds, which might be the reason that our pseudo label improvement is not as significant as our class localization maps.
A compatible solution for over-activation is expected in the future and it would potentially improve the segmentation results of \name even further.


\begin{table}[t!]
\caption{Evaluation of class localization maps on Deit-S backbones. 
aff: whether to use affinity refinement.
}
\centering\scalebox{.85}{
\setlength{\tabcolsep}{3mm}
\begin{tabular}{l c ccc}
\toprule
Method & Backbone & w/o aff & w/ aff \\ \hline
MCTformer~\cite{xu2022multi}& Deit-S  & 58.2 & 61.7  \\ 
MCTformer + ACR & Deit-S & 60.4 ($\uparrow$ 2.2)  & 62.4 ($\uparrow$ 0.7)  \\
\hline
\name & Deit-S & 56.8 & 63.4   \\

\hline

\end{tabular}}
\label{table ablation backbone}
\end{table}

\section{Conclusion}
In this paper, we propose a simple yet effective training framework to generate better class localization maps from transformer named \name. 
We exploit two types of consistencies during the classification training, i.e., 
region-wise activation consistency and region affinity consistency. 
The self-attention mechanism of the transformers is leveraged to simultaneously regularize two consistencies. 
We show that \name can learn precise object localization by only one single class token as well as 
accurate pair-wise affinity to extract object extent.
Our class localization maps significantly outperform previous methods and lead to state-of-the-art performances. 
\name can be seamlessly integrated with the vision transformer network without any extra modification, which can further facilitate other transformer-based tasks. 


{\small
\bibliographystyle{ieee_fullname}
\bibliography{egbib}
}

\appendix

\section{Implementation Details}
Our code is implemented in PyTorch on four NVIDIA RTX2080Ti GPUs.
During classification training, we use ViT-hybrid-B~\cite{dosovitskiy2020image} as the backbone.
The training images are randomly resized and cropped to 384 $\times$ 384 and we use a batch size of 4.  
The model is trained for 15 epochs using the SGD optimizer
with an initial learning rate of $0.01$, weight decay of $5e-4$, and Polynomial Learning Rate Policy.
In Equation 6, we set $\alpha=\beta=100$.


\paragraph{Evaluation Metric and Protocol}
For class localization maps, we report the best mean Intersection-over-Union (mIoU), i.e., the best match between the activation maps and the segmentation ground truth under all background thresholds.
For semantic segmentation (in mIoU), we obtain the PASCAL VOC \textit{val} and MS COCO results by comparing the predictions with their ground truth, while we obtain the PASCAL VOC \textit{test} results from the PASCAL VOC online evaluation server.

\section{Spatial Transformation and Inversion}

\begin{figure}[!t]
   \begin{center}
   {\includegraphics[width=1\linewidth]{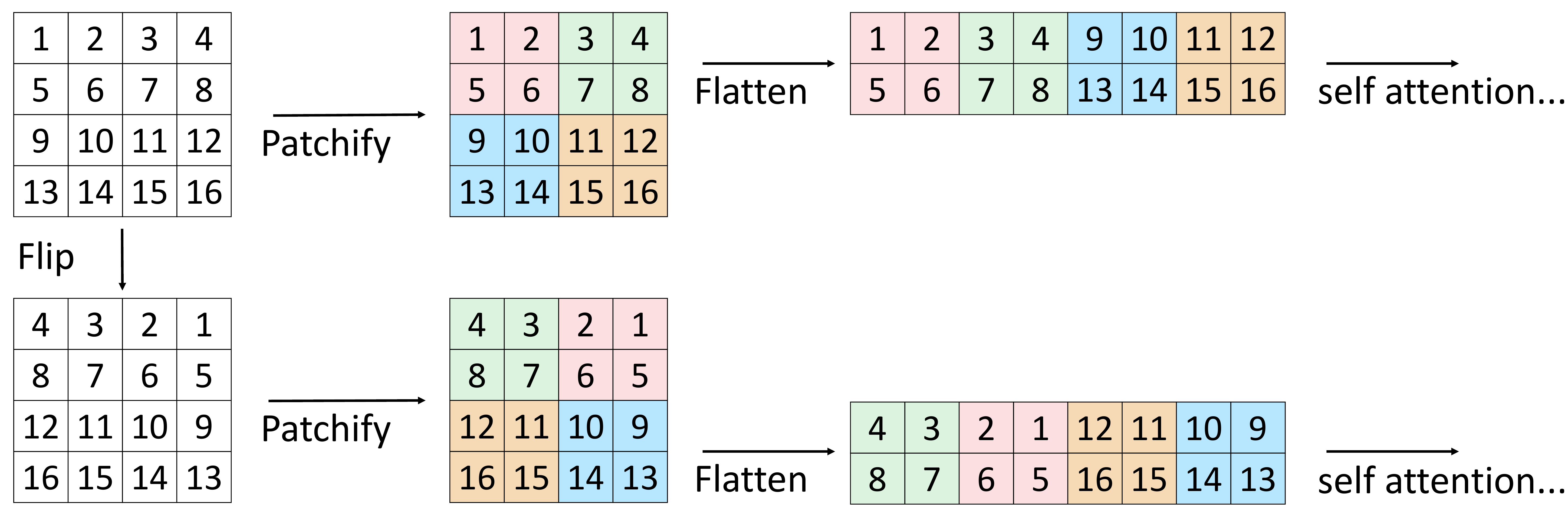}}
   \end{center}
\caption{Concept illustration of spatial transformation. The spatial augmentation not only transforms patch orders but also the content inside each patch. We propose transformation inversion to invert the patches to the original order.
}
   \label{fig:reverse}
\end{figure}

As discussed in the main paper Section 3.2, our consistency regularization requires an inverse transformation $f^{-1}$, which restores the spatial ordering of the tokens within the transformer. This is needed as the augmentation changes the pixel orderings in the spatial domain, which in turn alters the orderings of the patches, hence, tokens within the transformer (Fig.~\ref{fig:reverse}). Restoring the order is therefore necessary for us to match the corresponding patches \textit{before and after} the augmentation, so that we can compute losses.

Here, we present a toy example to demonstrate such an effect. As shown in Fig.~\ref{fig:reverse} (top), we have an input image of resolution $4 \times 4$ with unique number for each pixel. We then process it with a patch size of $2 \times 2$, with each patch in a different color. Then, we flatten the patches into a 1-d sequence of tokens as the input of the vision transformer.
In Fig.~\ref{fig:reverse} (bottom), we horizontally flip the image. 
Likewise, the augmented view is converted into patches that are flattened as a 1-d sequence of tokens. 
By comparing the colored blocks, a token-wise correspondence is easily drawn and order is restored. 
We point out that \textit{internally}, the embedding of a token is still different than its corresponding counterpart due to the augmentation applied in the pixel domain. But still, they shall share a similar activation signal as they are obtained from the same entity --- albeit flipped or rotated --- and our aim is to regularize such signals through losses.

To compute the losses for regularization, we require comparing elements \textit{inside} the self-attention matrices of the transformer, which means we need to restore the orderings of the said elements in a similar philosophy as what shown in Fig.~\ref{fig:reverse}. 
Though, this is far less trivial, as self-attention is calculated among all tokens. Below, we provide detailed derivations on how we obtain the inverse transformation $f^{-1}$ that restores such orderings.

\subsection{Derivation of the Transformation Inversion}
We consider the transformation inversion of the token ordering in this section.
By transformation inversion, we ensure the augmented attention matrix has an equal spatial ordering to the original attention matrix.
Note that we only consider the token ordering in this section and omit the transformation that has been applied inside each patch of the image as we only aim to restore the original spatial information, not the embedding.

\paragraph{Notations and Lemmas}

First, we define a general operation $\mathbf{vec}(\cdot)$ which converts an arbitrary 2-d vector (i.e., feature map) into a 1-d sequence
\begin{equation}
\mathbf{vec}: \mathbb R^{l\times m}\to \mathbb R^{lm \times 1},
\end{equation}
where $l$, $m$ denote any shape.
For an arbitrary ${H}=\left[h_{i j}\right]\in \mathbb R^{l\times m}$, $\mathbf{vec}(\cdot)$ yields its 1-d patched format, as
\begin{equation}
\mathbf{vec}(H)=\left[\begin{array}{c}
h_{11} \\
\vdots \\
h_{l 1} \\
\vdots \\
h_{1 m} \\
\vdots \\
h_{l m}
\end{array}\right]
\end{equation}
\noindent
In our setting, $h_{i j}$ represents the embedding derived from an image patch.

Second, we define a commutation matrix $C_{lm}\in \mathbb R^{lm\times lm}$, which fulfills
\begin{equation}
    \label{equation vec commutation lemma 2 }
    C_{lm} \mathbf{vec}({H})=\mathbf{vec}\left({H}^{{T}}\right).
\end{equation}
We can easily validate that ${C}_{{l} m}^{{T}}={C}_{l m}^{-1}={C}_{l m}$ ~\cite{petersen2008matrix}. Therefore, $C_{lm}$ is an orthogonal matrix. 

Finally, according to ~\cite{magnus1979commutation}, we have that for matrices
$A_{n\times p}, B_{p \times q}$, and  $C_{q \times m}$, the theorem holds that
\begin{equation}
    \label{equation vec commutation lemma}
    \mathbf{vec}({A B C})=\left({C}^{\mathrm{T}} \otimes A\right) \mathbf{vec}({B}),
\end{equation}
\noindent where $\otimes$ is Kronecker product. 

\paragraph{Spatial Transformation Operation}
Given an input image $I\in \mathbb R^{ H \times W}$ (channel dimension omitted for simplicity), 
we consider a spatial transformation operation (i.e., flipping or rotation) as a mapping of each individual pixel $(i,j) \in I$. Specifically, for flipping, we have three cases
\begin{eqnarray}
\text{horizontal flip: } (i, j)  &\to& (i, W-j), \nonumber \\
\text{vertical flip: } (i, j) &\to& (H-i, j), \nonumber \\
\text{horizontal \& vertical flip: } (i, j) &\to& (H-i, W-j), \nonumber
\end{eqnarray}
\noindent which can be represented by permutation operations.

Likewise, for rotation, we have
\begin{eqnarray}
\text{90}^{\circ} \text{rotate: } (i,j) & \to&  (W-j, H-i), \nonumber \\
\text{180}^{\circ} \text{rotate: } (i,j)& \to&   (j, H-i), \nonumber \\
\text{270}^{\circ} \text{rotate: } (i,j)& \to&  (W-j, I), \nonumber
\end{eqnarray}
\noindent while each case can be further considered as a matrix transpose followed by a flipping operation.

As such, we unify the above operations into matrix transformations. 
Given the feature map $X \in \mathbb{R}^{h \times w} $ of the said image $I$ obtained from e.g., ViT, where $h\times w = n$ ($n$ patches inside the transformer),
we have
\begin{eqnarray}
\text{flip: }  X  &\to& P_h X P_w, \label{flip} \\
\text{rotation: }  X  &\to& P_h X^T P_w, \label{rotate}
\end{eqnarray}
\noindent
where $P_h \in \mathbb{R}^{h \times h}$ and $P_w \in \mathbb{R}^{w \times w}$ are permutation matrices in the $x$ and $y$ directions  respectively.

\paragraph{Self-attention Matrices}
Here, we ask the question --- how will the self-attention matrices in the transformer change according to a spatial transformation operation on the image?

We assume $Q^{s}, K^{s}$ as the two projected feature maps of the input image, but in the 2-d shapes before flattening. 
Per the transformer attention design~\cite{vaswani2017attention}, we denote $Q^{s}=XW_Q$ and $K^{s}=XW_K$, where $Q^{s}, K^{s}$ are of dimension $\mathbb R^{h\times w\times d}$ with $d$ being the feature dimension. 
Here, $W_Q \text{ and } W_K$ project the embedded input image into two latent spaces, then we use $\mathbf{vec}(\cdot)$ to flatten $Q^s \text{ and } K^s$. 
The self-attention matrix of the original image before transformation is then defined as

\begin{equation}
A = \mathbf{vec}(Q^{s})(\mathbf{vec}(K^{s}))^T \in \mathbb{R}^{n \times n},
\end{equation}
\noindent
where we omit the class token for simplicity. 
Then, we write out the matrix after the transformation.
For flipping, the augmented self-attention matrix is formulated as
\begin{eqnarray}
\label{equation flip attention}
A' = (\mathbf{vec}(P_h Q^{s} P_w))(\mathbf{vec}(P_h K^{s} P_w))^T \in \mathbb{R}^{n \times n},
\end{eqnarray}
\noindent
which, per Equation~\ref{equation vec commutation lemma}, can be further derived as
\begin{eqnarray}
\label{eq inversion flip}
A' &=& (\mathbf{vec}(P_h Q^s P_w))(\mathbf{vec}(P_h K^s P_w))^T  \nonumber \\
&=& (P_w^T \otimes P_h)\mathbf{vec}(Q^s)((P_w^T \otimes P_h)\mathbf{vec}(K^s))^T \nonumber \\
&=& (P_w^T \otimes P_h)\mathbf{vec}(Q^s)\mathbf{vec}(K^s)^T(P_w^T \otimes P_h)^T \nonumber\\
&=& (P_w^T \otimes P_h)A(P_w^T \otimes P_h)^T .
\end{eqnarray}
\noindent


For rotation, the augmented self-attention is formulated as
\begin{equation}
\label{equation rotate attention}
A' = (\mathbf{vec}(P_h (Q^s)^T P_w))(\mathbf{vec}(P_h (K^s)^T P_w))^T \in \mathbb{R}^{n \times n} .
\end{equation}
Following the axiom of ~\ref{equation vec commutation lemma 2 } where $C \in \mathbb{R}^{n \times n}$ is an commutation matrix, Equation~\ref{equation rotate attention} can be rewritten as
\begin{eqnarray}
\label{eq inversion rotate}
A' &=& (\mathbf{vec}(P_h (Q^s)^T P_w))(\mathbf{vec}(P_h (K^s)^T P_w))^T  \nonumber \\
&=& (P_w^T \otimes P_h)C \mathbf{vec}(Q^s)((P_w^T \otimes P_h)C \mathbf{vec}(K^s))^T \nonumber \\
&=& (P_w^T \otimes P_h)C \mathbf{vec}(Q^s)\mathbf{vec}(K^s)^T C^T (P_w^T \otimes P_h)^T \nonumber\\
&=& (P_w^T \otimes P_h)CAC^T(P_w^T \otimes P_h)^T .
\end{eqnarray}
\noindent

\paragraph{Transformation Inversion}
At last, we obtain the formulation to invert the transformation on the attention matrices.
Following Equation~\ref{eq inversion flip} and Equation~\ref{eq inversion rotate}, the transformation inversion is in a  unified form
\begin{eqnarray}
\label{equation inverse}
f^{-1}(A') = C^{T}(P_w \otimes P_h^T){A'}{(P_w \otimes P_h^T)}^TC,
\end{eqnarray}
\noindent
where $f^{-1}$ is the inversion transformation. $C \in \mathbb{R}^{n \times n}$ is a commutation matrix for rotation and an identity matrix when flipping. 
Note that such a formulation
enables inversion of a wide range of possible image transformations that can be described with permutation matrices, though few may be helpful augmentations.
To this end,  $f^{-1}(A')$ and $A$ are spatially equivalent and 
we can directly calculate the distance between the two attention matrices to apply \name.

\begin{figure*}[!t]
   \begin{center}
   {\includegraphics[width=.9\linewidth]{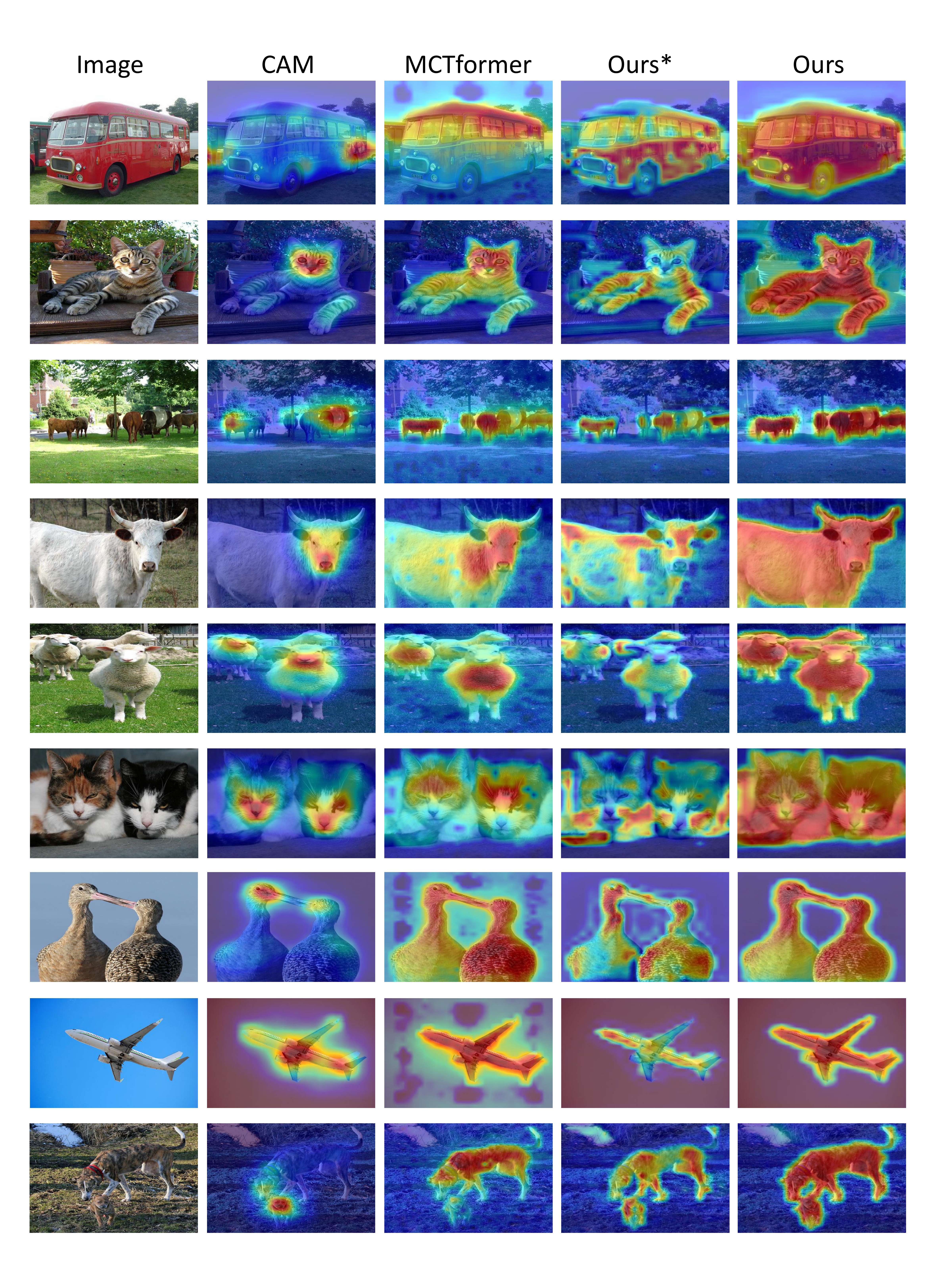}}
   \end{center}
   \vspace{-4mm}
\caption{Qualitative examples of class localization maps of \name. CAM:~\cite{zhou2016learning}. MCTformer:~\cite{xu2022multi}. Ours*: our maps without affinity refinement. Ours: our final class localization maps.}
   \label{fig:cam vis 2}
\end{figure*}

\begin{figure*}[!t]
   \begin{center}
   {\includegraphics[width=.9\linewidth]{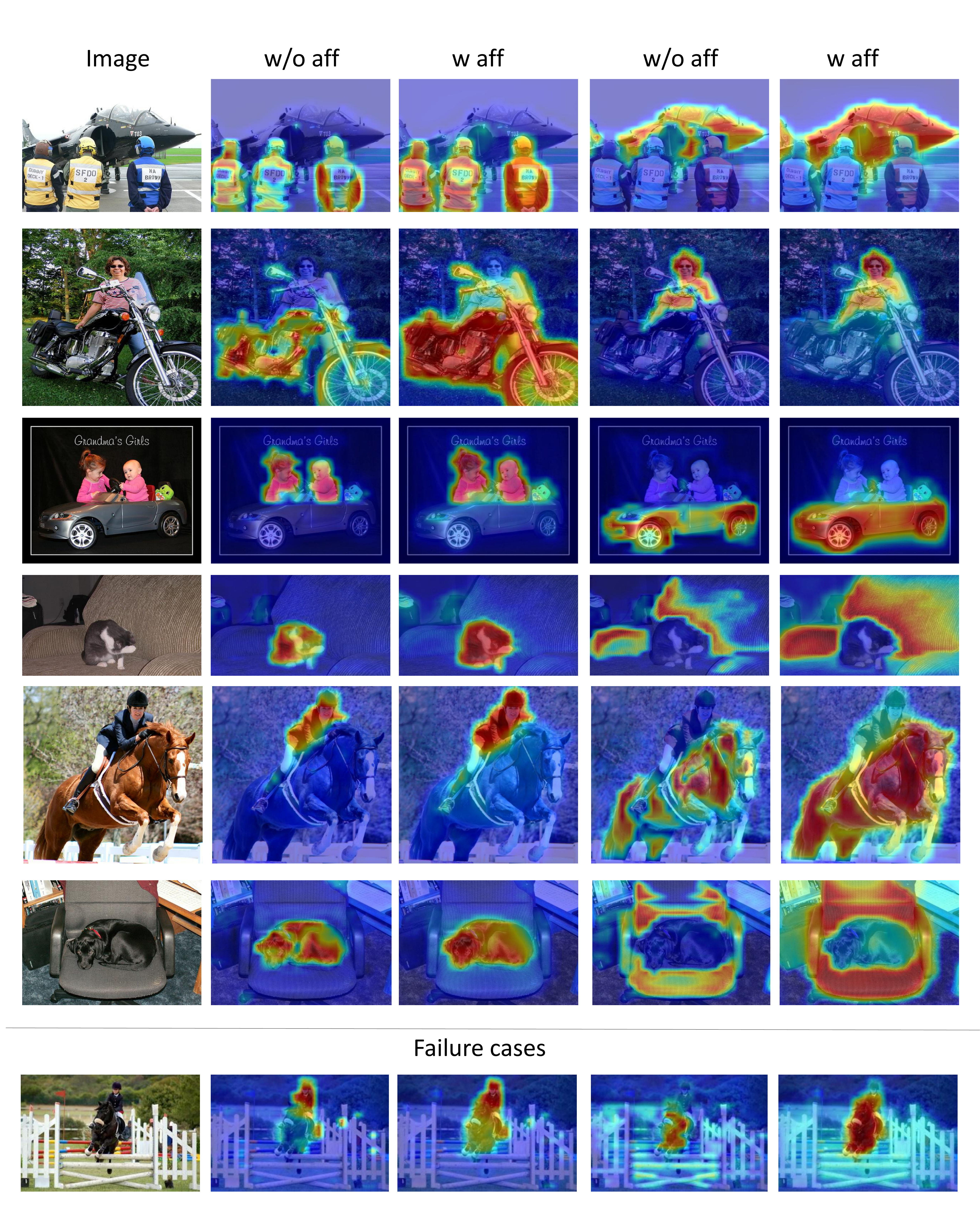}}
   \end{center}
\caption{Qualitative examples of our class localization maps with multiple classes. We show the results without and with affinity refinement. In the bottom, we present a failure case.}
   \label{fig:cam vis 3}
\end{figure*}

\begin{figure*}[!t]
   \begin{center}
   {\includegraphics[width=.6\linewidth]{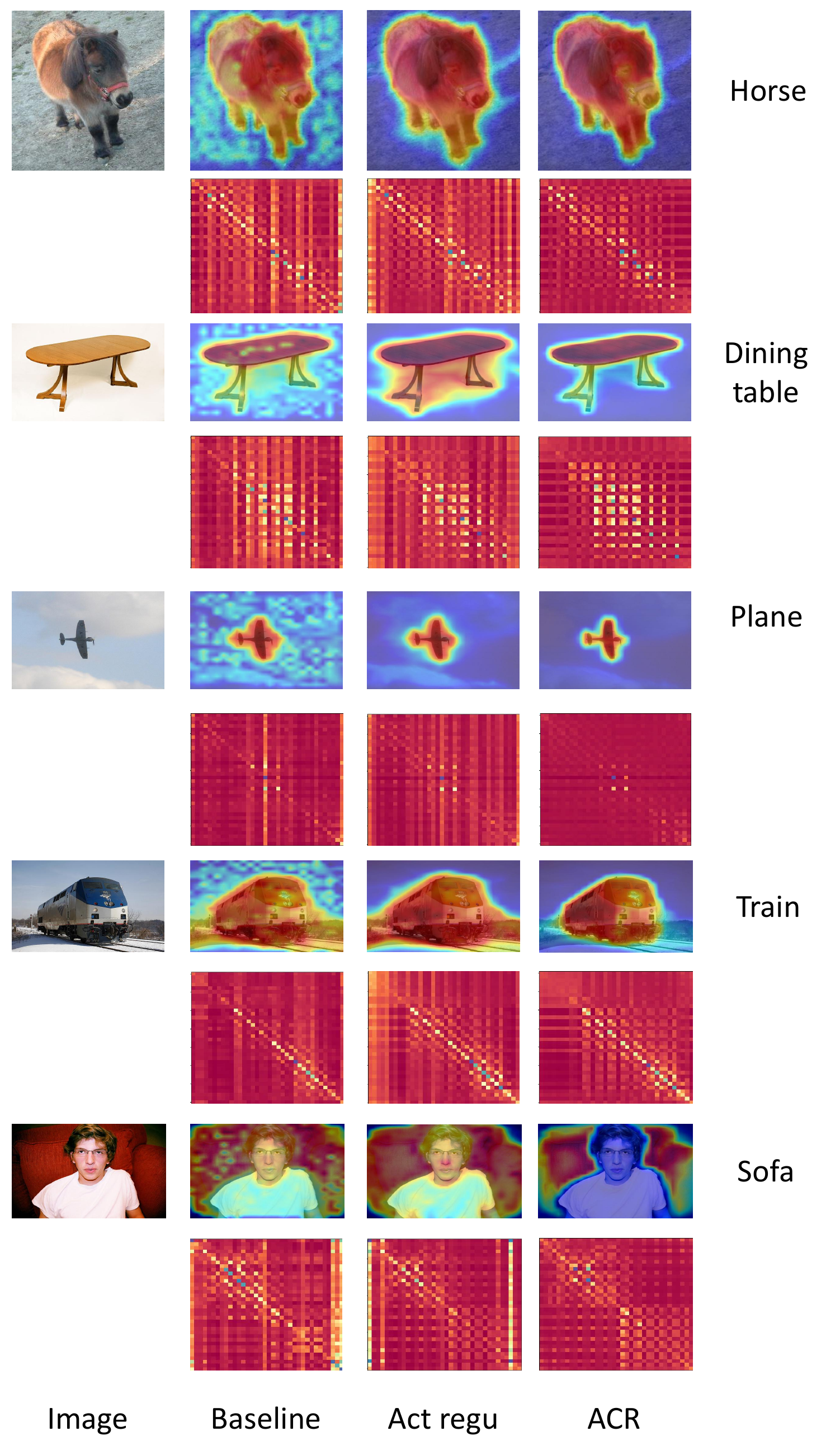}}
   \end{center}
\caption{Qualitative examples of the class localization maps and the learned affinity matrices. 
Baseline: the model is trained with only classification.
Act regu: the model is trained with only activation consistency regularization. ACR: the model is trained with our ACR that contains both consistency regularization. 
The baseline model with classification loss only generates noisy localization. 
With our activation regularization (Act regu), the model can correctly localize targeted objects but fails to capture precise boundaries.
Finally, our ACR can generate high-quality object localization maps, showing clearly the performance increase that arises from affinity consistency regularization.. The affinity matrices are down-sampled for readability.
}
   \label{fig:aff mat}
\end{figure*}

\begin{figure*}[!t]
   \begin{center}
   {\includegraphics[width=.9\linewidth]{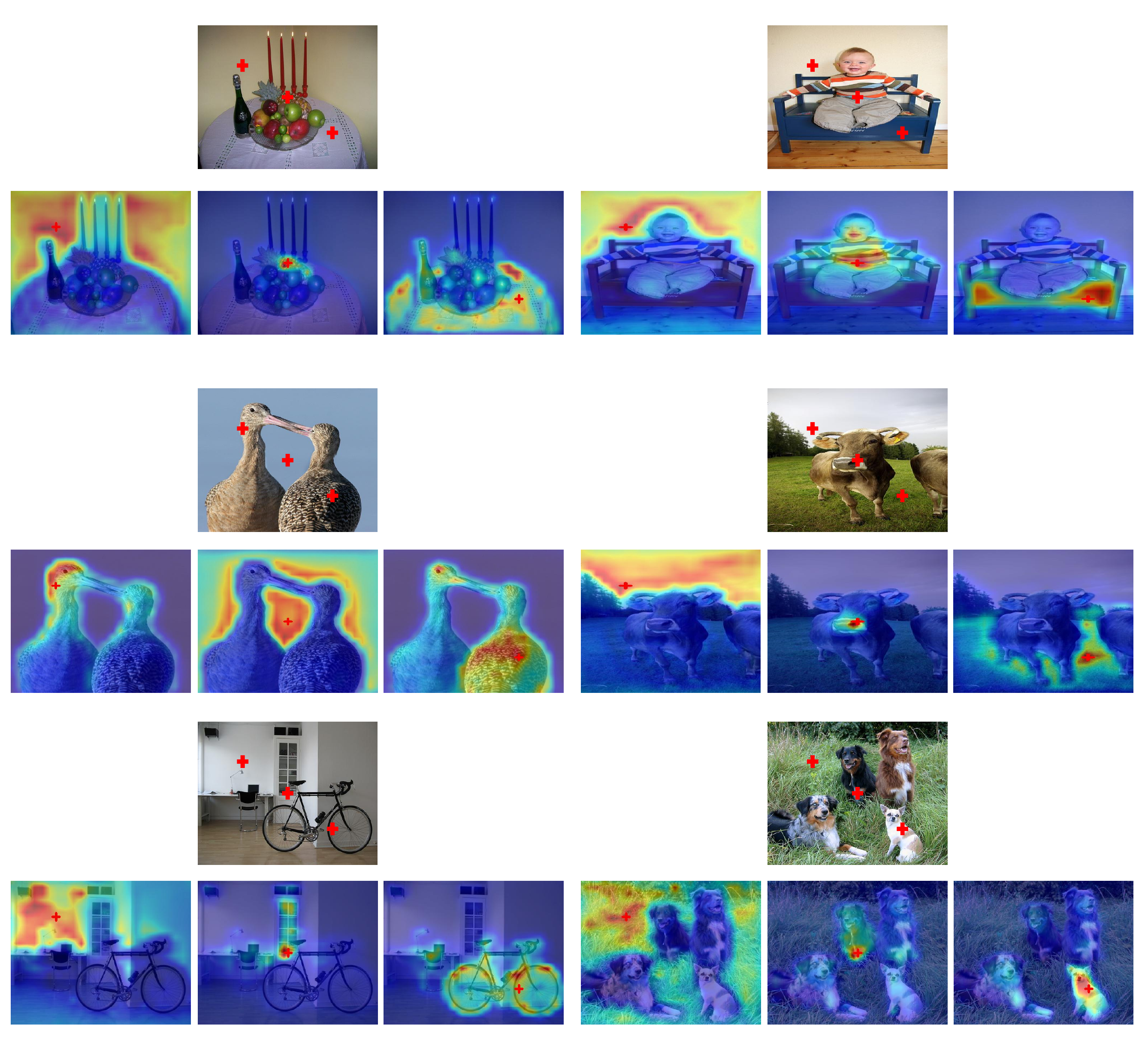}}
   \end{center}
\caption{Qualitative examples of the learned affinity of \name. Three source pixels are marked as red crosses. The region affinity related to these three source pixels is demonstrated respectively. Each source pixel is highly correlated with its semantically matched regions.}
   \label{fig:aff}
\end{figure*}

\begin{figure*}[!t]
   \begin{center}
   {\includegraphics[width=.9\linewidth]{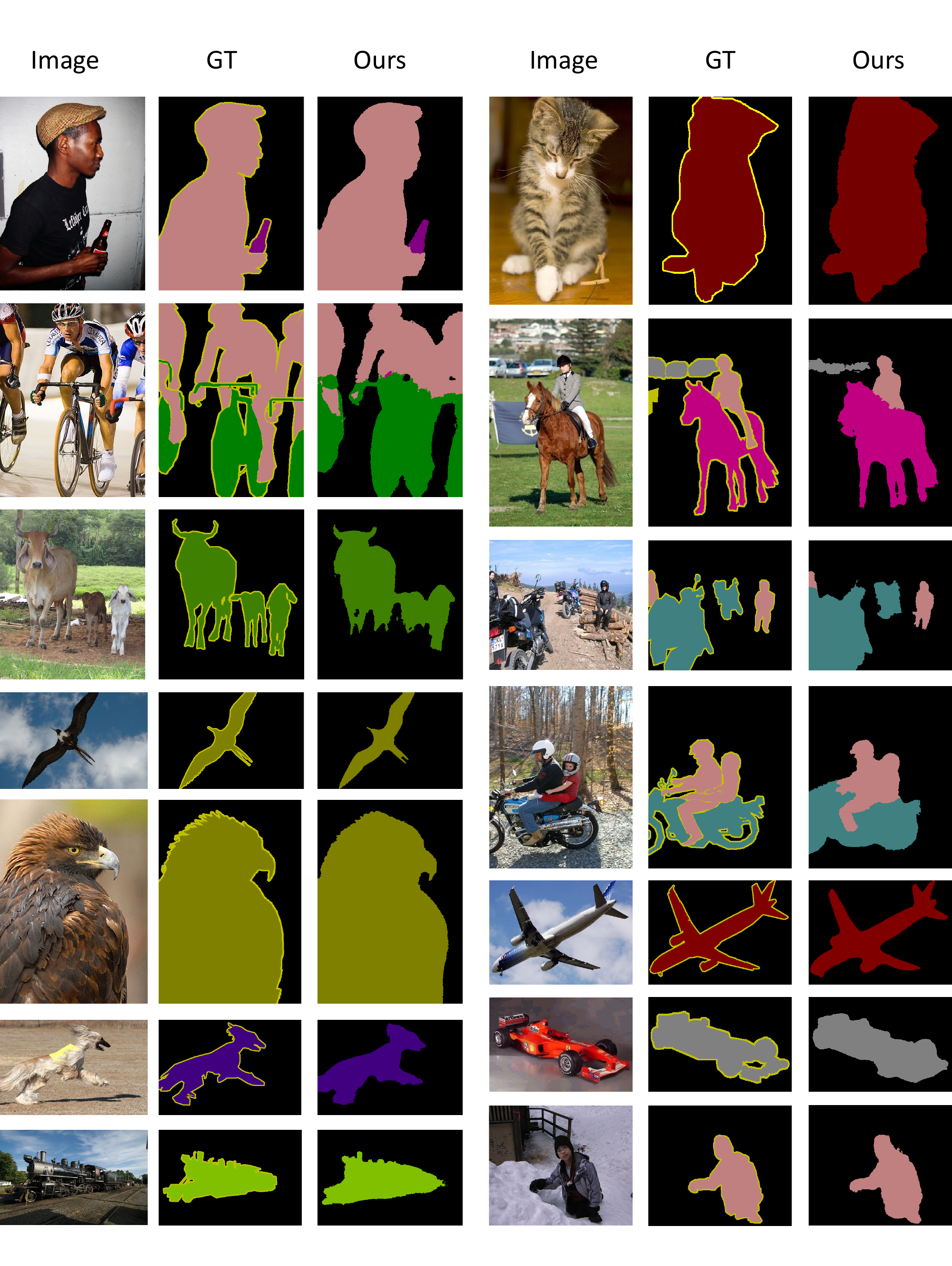}}
   \end{center}
\caption{Qualitative examples of the segmentation predictions}
   \label{fig:seg2}
\end{figure*}


\begin{table}[t!]
\centering
\caption{Ablation of different image augmentation methods. We report mIoU of seeds on PASCAL VOC \textit{train} set.}
\scalebox{.85}{
\setlength{\tabcolsep}{9mm}
\begin{tabular}{lc}
\toprule
Augmentation &   mIoU \\ \hline
Baseline (no aug) & 57.7 \\
Resize &  59.2 \\
Rotation & 61.1 \\
Horizontal flip + resize & 61.6  \\
Horizontal flip + vertical flip & 63.6  \\
Horizontal flip + patch hiding &  65.8   \\
Horizontal flip + gray scale  & 63.8   \\
Horizontal flip  &  67.3   \\
\bottomrule
\end{tabular}}
\label{table ablate augmentation}
\end{table}

\begin{table}[t!]
\centering
\caption{Ablation of different distance metrics for regularization loss. We report mIoU of seeds on PASCAL VOC \textit{train} set.}
\scalebox{.85}{
\setlength{\tabcolsep}{15mm}
\begin{tabular}{lc}
\toprule
Loss &   mIoU \\ \hline
L2 & 62.5  \\
Smooth L1 & 62.5  \\
L1 &  67.3   \\
\bottomrule
\end{tabular}}
\label{table ablate loss dist}
\end{table}

\begin{table}[t!]
\caption{Computational comparison. The training memory and test FPS are tested on an RTX2080Ti GPU with a batch size of 1.}
\centering\scalebox{.7}{
\begin{tabular}{l c c c c}
\toprule
Method & Backbone & Resolution & Train memory(MB) & Test FPS \\ \hline
PSA~\cite{ahn2018learning} & ResNet38 & 384 & 3082 & 0.98 \\
MCTformer~\cite{xu2022multi} & Deit-S & 224  & 1500 & 7.10 \\ 
Ours & Deit-S & 224 & 1580 & 3.61 \\
Ours & Vitb-hybrid-B & 384 & 5260 & 2.33 \\
\bottomrule
\end{tabular}}
\label{table: computation}
\end{table}

\section{Analysis of different image augmentation}
Several image augmentations are adopted in \name to transform the second view, 
we report the performance of them in Table~\ref{table ablate augmentation}.
As shown, we ablate several combinations of image augmentations. It is observed that all image augmentations achieve performance improvements over the baseline, which validates the effectiveness of \name.
Second, we found that horizontal flip on its own achieves the best result (67.3\% mIoU).
In future work, we will further investigate how different augmentations affect the performances. 


\section{More qualitative results}
We show more qualitative results of the class localization maps provided by \name. In Fig.~\ref{fig:cam vis 2}, we show class localization maps of images with simple scenes.
In Fig.~\ref{fig:cam vis 3}, we show that \name also generates high-quality class localization maps with multiple classes. 
In the bottom row of Fig.~\ref{fig:cam vis 3}, we show a failure case of an image containing a horse and a rider. 
Competitive relationships between the class activation are not investigated in this paper, so when we have multiple connected objects that have similar appearances or belong to the co-occurring classes, the affinity refinement may lead to over-activation. We will investigate this issue in future work.

Pair-wise relationships, or affinity, between image regions are inherently encoded in the attention matrix of the vision transformer.
The model is encouraged to capture consistent pair-wise affinity by our region affinity regularization.
We display the class localization maps and learned affinity matrices. in Fig.~\ref{fig:aff mat}.
The baseline model with classification loss only produces noisy localization maps, which is consistent with other methods, such as ~\cite{xu2022multi,sun2021getam,ru2022learning},
Further, if we only apply activation consistency regularization such as (ACT regu), model can correctly localize targeted objects but fails to capture precise object shapes.
As shown, some particular tokens are still causing the affinity matrices to become disorganized,
which indicates that simple activation consistency such SEAM~\cite{Wang_2018_CVPR} is not adequate and further affinity consistency is necessary. 
Finally, our ACR can generate high-quality object localization maps as the affinity regularization plays a key role in ensuring consistent appearance of object features, which in turn enhances segmentation performance
In addition, We show qualitative examples of the learned affinity in Fig.~\ref{fig:aff}.
We select three positions of the image which are marked as red crosses and show their related affinity. 
As shown, the learned affinity highly corresponds to semantic entities and shows accurate boundaries. For example, the background (wall, sky, ground) and foreground objects are clearly separated.
Such results indicate that our vision-transformer-based \name learns high-quality affinity and can effectively refine the class localization maps by propagating related pixels. 

Finally, we show qualitative examples of segmentation predictions in Fig.~\ref{fig:seg2}.

\section{Analysis of Regularization Loss}
Given two spatially equal attention matrices, we measure the distance between them. In Table~\ref{table ablate loss dist}, we ablate different types of distance evaluation methods and report the mIoU of the class localization maps on the PASCAL VOC \textit{train} set. As shown, we empirically found that L1 distance achieves the best result.

\section{Analysis of Efficiency}
In Table.~\ref{table: computation}, we compare our method with CNN-based PSA~\cite{ahn2018learning} and Transformer based MCTformer~\cite{xu2022multi}.
During training, our memory usages are similar under the same network and image size. 
Our inference speed is significantly faster than that of PSA.
However, our inference speed is slower than MCTformer due to the fact that MCTformer adopts multiple class tokens, whereas we maintain the model structure with a single class token and calculate gradients to generate class-wise localization.


\section{Limitations and Future Research}
We discuss the limitations and future research possibilities of our method in this section.
First, competitive relationships between class activation are not investigated in this paper, instead, our regularization is directly applied to class-indifferent attention matrices.
Thus, affinity refinement may lead to over-activation when multiple connected objects share similar appearances or belong to co-occurring classes. 
In future work, we will investigate how to connect the self-attention mechanism with the semantic relations between the classes so we can generate more class discriminative localization maps.
Second, as discussed in the main paper, 
our class localization maps are generally over-activated as the FP is consistently higher than the FN. It indicates that the incompleteness issue is effectively mitigated by \name.
However, current pseudo generation methods~\cite{ahn2018learning,ahn2019weakly}are designed for under-activated seeds, i.e., they require the seeds to have a high precision rather than recall.
It might be the reason that our pseudo label improvement is not as significant as our class localization maps.
A compatible solution for over-activation is expected in the future and it would potentially improve the segmentation results of \name even further.

\section{Per class results of PASCAL VOC and MS COCO}
We report the per class IoU of PASCAL VOC \textit{val} set and MS COCO \textit{val} set in Table~\ref{table per class pascal}  and Table~\ref{table ms coco per class}.

\begin{table*}[]
\centering
\caption{Per-class results on PASCAL VOC \textit{val} set.}
\scalebox{.8}{
\setlength{\tabcolsep}{1mm}
\begin{tabular}{l|cccccccccccccccccccccc}
\toprule
Class & bkg & plane & bike & bird & boat & bottle & bus & car & cat & chair & cow & table & dog & horse & motor & person & plant & sheep & sofa & train & tv & mIoU \\ \hline
CPN~\cite{Zhang_2021_ICCV}&
89.9& 75.1& 32.9& \textbf{87.8}& \textbf{60.9}& 69.5& \textbf{87.7}& 79.5& \textbf{89.0}& 28.0& 80.9& 34.8& 83.4& 79.7& 74.7& 66.9& 56.5& 82.7& 44.9& \textbf{73.1}& 45.7& 67.8 \\
Kweon et al.~\cite{Kweon_2021_ICCV} & 90.2 &82.9& 35.1 &86.8 &59.4 &70.6 &82.5& 78.1& 87.4& 30.1& 79.4& 45.9& 83.1& 83.4& \textbf{75.7}& 73.4& 48.1& \textbf{89.3}& 42.7& 60.4& 52.3& 68.4\\
Ours & \textbf{91.5} & \textbf{85.2} & \textbf{39.7} & 85.8 & 60.4 & \textbf{77.0} & 87.4 & \textbf{80.1} & 87.9 & \textbf{30.3} & \textbf{84.2} & \textbf{50.7} & \textbf{83.5} & \textbf{85.8} & 74.1 & \textbf{73.5} & \textbf{59.7} & 83.8 & \textbf{45.1}
 & 72.5 & \textbf{55.5} & \textbf{71.2} \\
\bottomrule
\end{tabular}}
\label{table per class pascal}
\end{table*}

\begin{table*}[]
\caption{Per-class results on MS COCO \textit{val} set.}
\setlength{\tabcolsep}{3.9mm}
\begin{tabular}{lccc|lccc}
\toprule
Class & MCTformer~\cite{xu2022multi} & RIB~\cite{lee2021reducing} & ours & Class & MCTformer~\cite{xu2022multi} & RIB~\cite{lee2021reducing} & ours \\ \hline
  background    & 82.4    &  82.0   & \textbf{82.7}     &   wine glass    &  27.0   & 27.5    & \textbf{48.2} \\
person    & \textbf{62.6}     &   56.1  &  47.0    &  cup    &  29.0    &   27.4  & \textbf{42.6} \\
  bicycle    &  47.4   &  \textbf{52.1}   & 50.4     &   fork   & 13.9    &\textbf{15.9}     & 12.6 \\
  car    &   \textbf{47.2}  &  43.6   &  44.6    &   knife    &  12.0   & 14.3    & \textbf{16.1} \\
  motorcycle    & 63.7    &  67.6   &  \textbf{68.4}    &   spoon    & 6.6    &  8.2   & \textbf{9.5} \\
airplane    &   64.7  &   61.3  &  \textbf{70.2}    &   bowl    &  22.4   &  20.7   & \textbf{26.5} \\
bus    &64.5     &   68.5  & \textbf{71.1}     &   banana    &  63.2   &  59.8   &  \textbf{64.3}\\
train    &  \textbf{64.5}   &  51.3   &  56.4    &   apple    & 44.4    &    \textbf{48.5} & \textbf{48.5} \\
  truck    &\textbf{44.8}     &   38.1  & 37.6     &   sandwich    &  39.7   &   36.9  & \textbf{51.0} \\
  boat    & \textbf{42.3}    & \textbf{42.3}    & 37.1     &   orange    & 63.0    &62.5     & \textbf{63.1} \\
  traffic light    &\textbf{49.9}     & 47.8    &  37.4    &   broccoli    & 51.2    &  45.4   & \textbf{53.8} \\
   fire hydrant   & 73.2    &   73.4  &  \textbf{74.9}    &   carrot    &  40.0   &     34.6& \textbf{44.3}\\
stop sign    &    \textbf{76.6} &   76.3  &  65.2    &   hot dog    &  \textbf{53.0}   &  49.7   & 52.1\\
parking meter    &   64.4  & \textbf{68.3}    &   50.8   &   pizza    & 62.2    &   58.9  & \textbf{79.3}\\
bench    &   32.8  &  39.7   &  \textbf{43.1}    &   donut    &  55.7   &  53.1   &\textbf{65.5} \\
bird    &   \textbf{62.6}  &  57.5   & 60.2     &   cake    &  47.9   & 40.7    & \textbf{52.6}\\
cat    &78.2     &  72.4   &  \textbf{78.4}    &   chair    & \textbf{22.8}    & 20.6    &18.7 \\
dog    &68.2     &  63.5   &  \textbf{72.0}    &   couch    & 35.0    &   36.8  &\textbf{39.9} \\
horse    &  65.8   &  63.6   &   \textbf{67.5}   &   potted plant    & 13.5    & 17.0    & \textbf{22.5}\\
sheep    &    70.1 &  69.1   &  \textbf{70.4}    &   bed    & 48.6    &  46.2   & \textbf{51.0}\\
cow    & 68.3    &  68.3   &  \textbf{71.4}    &   dining table    & 12.9    &  11.6   & \textbf{19.6}\\
elephant    &\textbf{81.6}     &  79.5   &  81.2    &   toilet    & 63.1    &   63.9  & \textbf{65.7}\\
bear    &   80.1  &  76.7   & \textbf{82.7}     &   tv    &  47.9   &   39.7  &\textbf{50.7} \\
zebra    &   \textbf{83.0}  &  80.2   &  82.1    &   laptop    &  49.5   &  48.2   &\textbf{54.6} \\
giraffe    &\textbf{76.9}     & 74.1    &  76.2    &   mouse    &  13.4   & \textbf{22.4}    &11.8 \\
backpack    &  14.6   &  \textbf{18.1}   &  13.3    &   remote    &  \textbf{41.9}   &  38.0   &37.4 \\
umbrella    &   61.7  &  60.1   &  \textbf{64.4}    &   keyboard    & 49.8    &  50.9   & \textbf{53.5}\\
handbag    &4.5     &  \textbf{8.6}   &  8.2    &   cellphone    & \textbf{54.1}    & \textbf{54.1}    & 53.2\\
tie    &    25.2 &  \textbf{28.6}   &  27.1    &   microwave    & 38.0    &  45.2   &\textbf{46.7} \\
suitcase    & 46.8    & \textbf{49.2}    &   48.3   &   oven    & 29.9    &  \textbf{35.9}   &32.7 \\
frisbee    &43.8     & 53.6    & \textbf{57.0}     &   toaster    & 0.0    &\textbf{17.8}     &0.0 \\
skis    &   12.8  & 9.7    &   \textbf{14.1}   &   sink    & 28.0    &   \textbf{33.0}  &30.4 \\
snowboard    &  \textbf{31.4}   &  29.4   &  23.7    &   refrigerator    &  40.1   & \textbf{46.0}    &32.9 \\
sports ball    &    9.2 &  \textbf{38.0}   & 21.5     &   book    &  32.2   &   31.1  &\textbf{33.2} \\
kite    & 26.3    &  37.1   & \textbf{47.1}     &   clock    & 43.2    &    41.9 & \textbf{52.6}\\
baseball bat    & 0.9    &  \textbf{15.3}   &   11.0   &   vase    &  22.6   &    27.5 & \textbf{31.4}\\
baseball glove    &     0.7& \textbf{8.1}    &  7.1    &   scissors    &  32.9   &   41.0  & \textbf{42.4}\\
skateboard    & 7.8    &  \textbf{31.8}   &  26.0    &   teddy bear    & 61.9    &   \textbf{62.0}  & 60.3\\
surfboard    &   \textbf{46.5}  &  29.2   &  38.6    &   hair drier    & 0.0    &   \textbf{16.7}  & 0.0\\
tennis racket    &  1.4   &   \textbf{48.9}  &   21.0   &   toothbrush    & 11.1    &   21.0  & \textbf{31.0}\\
bottle    &31.1     &  33.1   &  \textbf{38.5}    &   mIoU    & 42.0    &   43.8  & \textbf{45.0} \\
\bottomrule
\end{tabular}
\label{table ms coco per class}
\end{table*}

\end{document}